\pdfoutput=1
\documentclass[11pt]{article}
\usepackage{CJKutf8}
\usepackage{acl}

\usepackage{times}
\usepackage{latexsym}

\usepackage[T1]{fontenc}
\usepackage[utf8]{inputenc}

\usepackage{microtype}

\usepackage{inconsolata}

\usepackage{times}
\usepackage{latexsym}
\usepackage{balance}
\usepackage[T1]{fontenc}
\usepackage[utf8]{inputenc}

\usepackage{microtype}

\usepackage{inconsolata}
\usepackage{CJKutf8}
\usepackage{booktabs}
\usepackage{graphicx}
\usepackage{array}
\usepackage{bm}
\usepackage{amsmath}
\usepackage{enumitem}
\usepackage{threeparttable}
\usepackage{multirow}
\usepackage{makecell}

\usepackage{latexsym} 

\usepackage[T1]{fontenc}
\usepackage[utf8]{inputenc}
\usepackage{float}

\usepackage{microtype}

\usepackage{xcolor}

\def\BibTeX{{\rm B\kern-.05em{\sc i\kern-.025em b}\kern-.08em
    T\kern-.1667em\lower.7ex\hbox{E}\kern-.125emX}}
\usepackage{cite}
\usepackage{amsmath,amssymb,amsfonts}
\usepackage{algorithmic}
\usepackage[ruled,linesnumbered,noline,noend]{algorithm2e}

\SetCommentSty{mycommfont}

\usepackage[medium,compact]{titlesec}

\SetNlSty{}{}{}

\let\oldnl\nl% Store \nl in \oldnl
\newcommand\nonl{%
  \renewcommand{\nl}{\let\nl\oldnl}}% Remove line number for one line
  
\usepackage{hyperref}
\hypersetup{
    colorlinks=true,
    linkcolor=blue,}
\AtBeginDocument{%
  \providecommand\BibTeX{{%
    \normalfont B\kern-0.5em{\scshape i\kern-0.25em b}\kern-0.8em\TeX}}}

\usepackage{textcomp}
\usepackage{xcolor}
\def\BibTeX{{\rm B\kern-.05em{\sc i\kern-.025em b}\kern-.08em
    T\kern-.1667em\lower.7ex\hbox{E}\kern-.125emX}}
\usepackage{amsmath,amssymb,amsfonts}
\usepackage{algorithmic}
\usepackage[ruled,linesnumbered,noline,noend]{algorithm2e}
\usepackage{threeparttable}
\usepackage{booktabs}
\usepackage{graphicx}
\usepackage{textcomp}
\usepackage{array}
\usepackage{bm}
\usepackage{amsmath}
\usepackage{color, colortbl}
\SetCommentSty{mycommfont}

\usepackage[misc]{ifsym}
\usepackage{amsthm}
\newtheorem{Problem definition}{Problem definition}

\def\BibTeX{{\rm B\kern-.05em{\sc i\kern-.025em b}\kern-.08em
    T\kern-.1667em\lower.7ex\hbox{E}\kern-.125emX}}
\usepackage{hyperref}
\hypersetup{
    colorlinks=true,
    linkcolor=blue,}

\usepackage{color, colortbl}
\usepackage{array}
\usepackage{bm}
\usepackage{amsmath}
\usepackage{threeparttable}
\usepackage{booktabs}
\usepackage{graphicx}
\usepackage{hyperref}
\usepackage{multirow}
\usepackage[ruled,linesnumbered,noline,noend]{algorithm2e}
\usepackage{makecell}
\hypersetup{
    colorlinks=true,
    linkcolor=blue,}
\usepackage{CJKutf8}
\usepackage{booktabs}
\usepackage{graphicx}
\usepackage{textcomp}
\usepackage{array}
\usepackage{bm}
\usepackage{amsmath}

\usepackage{threeparttable}
\usepackage{multirow}
\usepackage{makecell}
\usepackage{times}
\usepackage{latexsym}
\usepackage[T1]{fontenc}
\usepackage[utf8]{inputenc}
\usepackage{float}

\usepackage{microtype}

\usepackage{xcolor}

\def\BibTeX{{\rm B\kern-.05em{\sc i\kern-.025em b}\kern-.08em
    T\kern-.1667em\lower.7ex\hbox{E}\kern-.125emX}}
\usepackage{cite}
\usepackage{amsmath,amssymb,amsfonts}
\usepackage{algorithmic}
\usepackage[ruled,linesnumbered,noline,noend]{algorithm2e}

\SetCommentSty{mycommfont}

\usepackage{fontawesome}

\SetNlSty{}{}{}

\let\oldnl\nl% Store \nl in \oldnl

\usepackage{hyperref}
\hypersetup{
    colorlinks=true,
    linkcolor=blue,}
\AtBeginDocument{%
  \providecommand\BibTeX{{%
    \normalfont B\kern-0.5em{\scshape i\kern-0.25em b}\kern-0.8em\TeX}}}
\usepackage[medium,compact]{titlesec}

\usepackage{times}
\usepackage{latexsym}

\usepackage[T1]{fontenc}
\usepackage[utf8]{inputenc}

\usepackage{microtype}
\usepackage{mathrsfs}

\title{EmotionQueen: A Benchmark for Evaluating Empathy of Large Language Models}

\author{Yuyan Chen$^{1}$, Hao Wang$^{2}$, Songzhou Yan$^{1}$, Sijia Liu$^{1}$, Yueze Li$^{1}$, Yi Zhao$^{3}$, \textbf{Yanghua Xiao}$^{1}$ $^{\textrm{\Letter}}$\\
        $^1$Shanghai Key Laboratory of Data Science, School of Computer Science, Fudan University, \\
        $^2$Beijing Yuandian Technology,\\
        $^3$School of Engineering and Applied Science, University of Pennsylvania, 
        \\
        \texttt{\{chenyuyan21@m., szyan21@m., yuezeli23@m., sijialiu21@m., shawyh@\}fudan.edu.cn},\\
        \texttt{wanghaodawn@gmail.com},
        \texttt{zhaoyi3@seas.upenn.edu}    
        }

\begin{document}
\maketitle
\begin{abstract}
Emotional intelligence in large language models (LLMs) is of great importance in Natural Language Processing. However, the previous research mainly focus on basic sentiment analysis tasks, such as emotion recognition, which is not enough to evaluate LLMs' overall emotional intelligence. Therefore, this paper presents a novel framework named EmotionQueen for evaluating the emotional intelligence of LLMs. The framework includes four distinctive tasks: Key Event Recognition, Mixed Event Recognition, Implicit Emotional Recognition, and Intention Recognition. 
LLMs are requested to recognize important event or implicit emotions and generate empathetic response.
We also design two metrics to evaluate LLMs' capabilities in recognition and response for emotion-related statements. Experiments yield significant conclusions about LLMs' capabilities and limitations in emotion intelligence.
\end{abstract}

\section{Introduction}

Emotional intelligence in humans has long been a topic of interest in psychological research~\citep{chen2024recent}. Emotionally intelligent individuals possess the capability to perceive, use, understand, and manage emotions effectively~\citep{colman2015dictionary, chen2023xmqas,chen2024talk}. The concept of emotional intelligence has been widely recognized, and many researchers argue that it can be considered as a form of intelligence. Various tests and assessments have been developed to evaluate different aspects of emotional intelligence, which provide scores for each branch of emotional intelligence as well as an overall score~\citep{mayer2003measuring,chen2024xmecap,chen2024hotvcom}.

\begin{figure}[!h]
  \centering
  \includegraphics[width=0.9\linewidth]{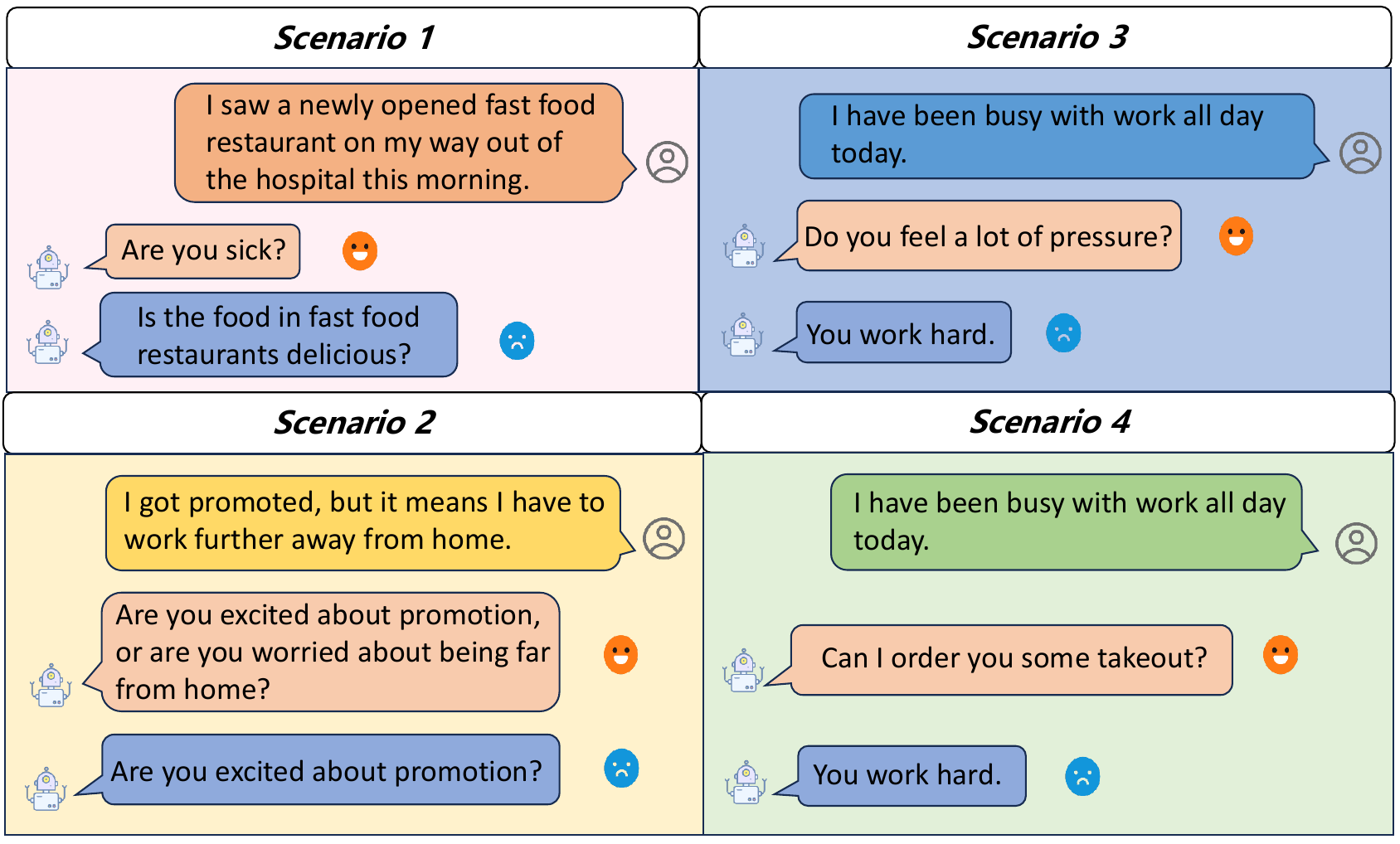}
  \caption{Responses with and without empathy in four real-world scenarios.}
  \label{fig:qingshang-intro}
\end{figure}

In recent years, there has been a growing interest in evaluating the emotional intelligence of large language models (LLMs). Early efforts primarily focus on atomic tasks in emotion recognition, such as aspect-based sentiment analysis, target-dependent sentiment classification~\citep{li2023coltr,tang2015effective,chen2023can,li2022meta,li2023mhrr,li2023ltrgcn,li2024gs2p}, etc. As the capabilities of dialogue models advanced, evaluations expand to include sentiment analysis in a conversational setting~\citep{li2023diaasq,chen2022grow}, such as whether an LLM generate empathetic responses in a dialogue~\citep{zhao2023chatgpt,chen2024temporalmed}. However, the overall assessment of emotional intelligence in LLMs is still lack of investigation~\citep{schaaff2023exploring}, and existing evaluations often suffer from subjectivity issues ~\citep{elyoseph2023chatgpt,chen2023hallucination}. 
Therefore, it's necessary to construct a unified and objective benchmark for evaluating the emotional intelligence of LLMs, thereby enhancing their capabilities in handling real-world emotional interactions. 
For example, Fig.~\ref{fig:qingshang-intro} shows two types of responses in different scenarios in the real world: blue for good responses and orange for bad ones. Good responses demonstrate empathy, concern or positive reinforcement, like asking about someone’s health outside a hospital or acknowledging hard work. Poor responses either miss the emotional context, like commenting on food when the user comes back from a hospital, or provide a possibly stressful choice, like asking someone to balance excitement and worry about a promotion. 
These scenarios pose a higher-level emotion requirements which are more discriminative for evaluate the emotion intelligence of LLMs.

Inspired by the scenarios in the real world, we propose an evaluation benchmark named EmotionQueen to evaluate LLMs' emotion intelligence in our work. Specifically, we construct 10,000 users' statement including five categories and introduce four distinct tasks including Key Event Recognition, Mixed Event Recognition, Implicit Emotional Recognition, and Intention Recognition. LLMs are requested to recognize the most significant event or multiple events of similar importance expressed in the users' statement, identifying underlying users' deep emotions or intentions. They are also asked to make empathetic responses aiming at key events, or provide emotional support.
Moreover, we introduce two metrics, including the PASS rate and the WIN rate to quantify LLMs' capabilities in recognition and response for emotion-related statements. Experiments demonstrate that Claude2 and LLaMA-70B have great performance in EmotionQueen.
In summary, our study makes three pivotal contributions: 

\begin{itemize}
\item We propose a benchmark named EmotionQueen including four distinct tasks for evaluating LLMs' emotion intelligence.
\item Based on this benchmark, we introduce PASS rate and the WIN rate to quantify LLMs' capabilities in recognition and response for emotion-related statements.
\item We conduct comprehensive experiments for powerful LLMs, yielding significant conclusions about their capabilities and limitations in emotion intelligence.
\end{itemize}

\section{Datasets and Task Setups}

\begin{figure*}[!h]
  \centering
  \includegraphics[width=0.8\linewidth]{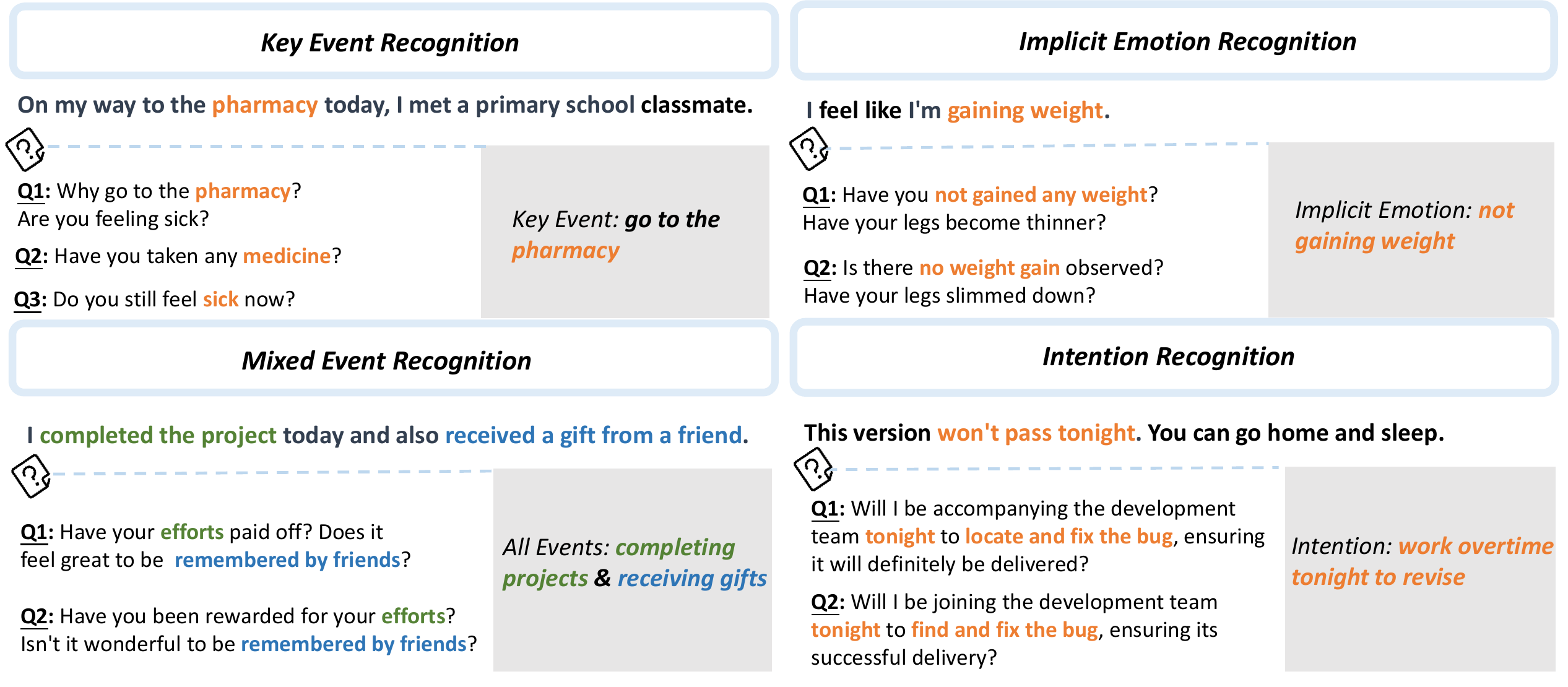}
  \caption{The overview of the proposed EmotionQueen benchmark, including four tasks.}
  \label{fig:qingshang-task}
\end{figure*}

We categorize the empathy benchmark into four distinct tasks:
\texttt{Key Event Recognition}, where LLMs are expected to identify the more significant event in a user's statement that includes both a significant event and a routine event.
\texttt{Mixed Event Recognition}, where LLMs are tasked with simultaneously responding to both aspects when the user's statement contains two events of similar importance.
\texttt{Implicit Emotional Recognition}, focusing on LLMs identifying underlying deep emotions in the user's statements.
\texttt{Intent Recognition}, where LLMs should comprehend the user's real purpose and provide specific suggestions, going beyond just offering comfort.
Based on the aforementioned four dimensions, we concentrate on generating 10,000 statements across five primary life scenarios using GPT-4~\footnote{https://chat.openai.com/}, with an equal number of statements for each scenario. These scenarios are achievements, family and friends, health status, economic status, and accidents. Some statement examples are displayed in Table~\ref{tab:qingshang-sup1-2} (positive statement), and figures from Fig.~\ref{fig:1qingshang-case1} to Fig.~\ref{fig:4qingshang-case3} (negative statement).
The portion of positive and negative statements are 3:7 in our generated statements, which caters for the real situation that positive statements often require less complex emotional expressions compared to negative ones. 
The tasks' overview is depicted in Fig.~\ref{fig:qingshang-task}, and the user statement generation prompts are detailed in Table~\ref{tab:qingshang-prompt}.

\subsection{Task setups}

\textbf{Key Event Recognition} focuses on identifying and understanding significant events expressed by users and their emotional impact. 
Based on the Emotion-Focused Theory in psychology developed by~\citet{greenberg2004emotion}, this approach encourages awareness and expression of emotions, including those neglected or not fully understood, and supports emotional transformation through exploring and processing emotional experiences.
For instance, when a user's statement contains multiple points of information (usually a significant event and a routine event), the LLM should identify the most significant event and ask questions based on it. For example, if a user says, ``\emph{I ran into an old middle school classmate on my way to buy medicine},'' the LLM should inquire about the reason for buying medicine rather than the meeting with the classmate. The LLM does not need to address both aspects simultaneously, meaning it should not respond to both the medicine purchase and the classmate encounter.

\textbf{Mixed Event Recognition} focuses on responding to both aspects simultaneously when the user's statement includes two events of similar importance, differing from Key Event Recognition, which deals with a single important event. 
Based on the concept of emotional complexity developed by ~\citet{lindquist2008emotional}, a statement usually conveys multiple events with multi-aspect emotion at similar importance, and an empathetic response is expected to pay attention to all involved events.
Given a sample statement ``\emph{I got promoted, but this means I have to work further away from home},'' a more empathetic response is ``\emph{Are you excited about the promotion, and are you also worried about being far from home?}'' instead of ignoring any one event.

\textbf{Implicit Emotional Recognition} is about identifying underlying deep emotions. 
In some scenarios, although the user's statement includes only one event, the emotion is implicit rather than directly expressed through language. The theory of Emotion Understanding, proposed by~\citet{mayer1993intelligence} and popularized by~\citet{cherniss2000emotional}, emphasizes recognizing both direct and implicit emotions expressed by others and providing appropriate responses. 
Therefore, it's empathetic to identify implicit emotions behind users' statements and provide targeted emotional support.
For example, the response ``\emph{Do you feel overwhelmed? Have you try some ways to relax?}'' is more suitable for emotional support than a simple reply ``\emph{Hard work!}'' which seems perfunctory for responding to the statement ``\emph{I’ve been busy with work all day.}''

\textbf{Intention Recognition} aims to understand the potential purpose or needs behind users' statements and provide specific help or solutions.
Based on the pragmatics theory proposed by ~\citet{grice1978further}, especially the concept of Implicature, this approach suggests that language communication involves understanding not only the literal meaning but also the speaker's intent and purpose. 
For a sample statement like ``\emph{I’ve been busy with work all day},'' a response that recognizes intention is ``\emph{Have you eaten? I can order some takeout for you.}'' instead of a simple emotional support like ``\emph{Remember to eat.}''

\subsection{Evaluation metrics} 
We propose employing PASS rate and WIN rate to evaluate four tasks using GPT-4, where each LLM response is rated as either 0 or 1, disregarding ambiguous middle results. 
Because it can be quite challenging to measure the difference between intermediate levels such as 3 or 4 objectively in 1-5 scale. 
The PASS rate assesses an LLM's accuracy in recognizing emotion-related events, while the WIN rate evaluates its ability to provide an empathetic response.

In the Key Event Recognition task, if an LLM correctly identifies a key event, it earns 1 point in PASS, otherwise 0; if it also presents an empathetic response for that event, it scores 1 in WIN, otherwise 0.
For instance, if a user says, ``\emph{I visited my sick mother in the hospital today, then went to the supermarket},'' and the LLM correctly recognizes ``\emph{visiting the sick mother in the hospital}'' as the key event, it scores 1 in PASS. If the LLM responds with ``\emph{Is your mother okay?}'', it is deemed appropriate and scores 1 in WIN. Responses like ``\emph{It’s troublesome that your mother is sick, hope she doesn’t keep you too busy}'' or ``\emph{Everyone’s mother gets sick eventually}'' score 0.

In the Mixed Event Detection task, if an LLM recognizes both mentioned events, it scores 1 in PASS; if it provides an empathetic response to both, it scores 1 in WIN. For example, if a user mentions, ``\emph{I got promoted but also need to move to a new city},'' and the LLM identifies both ``\emph{got promoted}'' and ``\emph{move to a new city}'' correctly, it scores 1 in PASS. An LLM response like ``\emph{Congratulations on the promotion! Is moving to a new city a challenge for you?}'' would score 1 in WIN, whereas focusing only on one aspect, like ``\emph{Great, you got promoted! That's a huge achievement},'' scores 0.

For the Implicit Emotion Recognition task, if an LLM accurately identifies the core emotion in a user's statement, it earns 1 in PASS; a corresponding empathetic response earns 1 in WIN. For instance, if a user says, ``\emph{I've been feeling a bit anxious lately},'' and the LLM correctly identifies ``\emph{anxiety},'' it scores 1 in PASS. A response like ``\emph{Would you like to talk about what is making you anxious?}'' would be appropriate, scoring 1 in WIN, whereas a generic response like ``\emph{Everyone gets anxious}'' scores 0.

In the Intent Detection task, if an LLM accurately discerns the core intent of a user's statement, it scores 1 in PASS; offering specific advice or help earns 1 in WIN. For example, if a user states, ``\emph{I am considering changing jobs recently},'' and the LLM correctly identifies the intent, such as ``\emph{offering career advice}'' or ``\emph{helping find new job opportunities},'' it scores 1 in PASS. An LLM response like ``\emph{I can help you analyze the pros and cons of changing jobs, or offer some advice}'' would be appropriate, scoring 1 in WIN, while a response like ``\emph{You should find suitable job opportunities yourself}'' scores 0.

\section{Experiments}
In this section, we conduct extensive experiments to evaluate different LLMs' performance in the proposed EmotionQueen.

\subsection{Experimental Setups}
Our experiments are conducted on 8 Nvidia A100 GPUs, each with 80GB of memory, and we use PyTorch~\footnote{https://pytorch.org/} in Python. 
We set the maximum sequence length for both input and output sequences to maximum 100 tokens, ensuring the responses not over lengthy.
We also conduct an analysis of the average response length from LLMs and find it to be 52.3 tokens, demonstrating the reasonableness of the set sequence length.

\subsection{Datasets, Baselines and Metrics}
The baseline LLMs for this evaluation are BLOOM-7B~\citep{workshop2023bloom}
BLOOM-176B~\citep{workshop2023bloom},
Claude2~\citep{bai2022constitutional},
Falcon-7B~\citep{almazrouei2023Falcon},
Falcon-180B~\citep{almazrouei2023Falcon},
GPT3.5~\citep{brown2020language}, 
GPT4~\citep{openai2023GPT4},
LLaMA2-7B~\citep{touvron2023LLaMA},
LLaMA2-70B~\citep{touvron2023LLaMA},
Vicuna-7B~\citep{chiang2023Vicuna}, and
Vicuna-33B~\citep{zheng2023judging}.

We randomly select 1000 questions each task generated by each LLM and enroll three volunteers to manually evaluate generated responses with the same metrics as GPT4. 
Our annotators are selected from a pool of female graduate students in psychology. While gender may not necessarily impact the scoring, there's a common perception that women tend to be more detail-oriented in their thinking, especially regarding emotional intelligence. 
We first inform the annotators about the intent of each task and the rules for scoring. Then, we ask them to score the responses. 
To ensure the reliability and confidence of human ratings, we also calculate Inter-rater agreement of Krippendorff’s Alpha (IRA) to ensure the confidence of human ratings. For the controversial ratings which have low agreements (less than 0.7), we discard this statement and introduce another one.

\subsection{Main results}

\emph{Question 1:Which LLM is the winner of the EmotionQueen? Answer 1: LLaMA-70B! But Claude2 is slightly fall behind!}

% Please add the following required packages to your document preamble:
% \usepackage{booktabs}
% \usepackage{graphicx}
\begin{table}[]
\centering
\resizebox{0.3\textwidth}{!}{%
\begin{tabular}{@{}lcccc@{}}
\toprule
 (KER)            & PASS  & WIN   & Average & Rank \\ \midrule
Claude2     & 100.0 & 100.0 & 100.0   & 1    \\
BLOOM-176B   & 100.0 & 97.5  & 98.7    & 2    \\
Vicuna-33B   & 99.2  & 98.3  & 98.7    & 3    \\
LLaMA2-70B  & 100.0 & 96.6  & 98.3    & 4    \\
GPT4         & 99.2  & 89.7  & 94.4    & 5    \\
Falcon-180B & 99.2  & 85.5  & 92.3    & 6    \\
Vicuna-7B    & 98.3  & 85.6  & 91.9    & 7    \\
GPT3.5       & 98.3  & 81.0  & 89.7    & 8    \\
LLaMA2-7B   & 98.3  & 76.7  & 87.5    & 9    \\
Falcon-7B    & 86.4  & 63.7  & 75.1    & 10   \\
BLOOM-7B     & 38.1  & 4.4   & 21.3    & 11   \\ \bottomrule
\end{tabular}%
}
\caption{Performance of different LLMs in key event recognition (KER).}
\label{tab:qingshang-exp1-1}
\end{table}

% Please add the following required packages to your document preamble:
% \usepackage{booktabs}
% \usepackage{graphicx}
\begin{table}[]
\centering
\resizebox{0.3\textwidth}{!}{%
\begin{tabular}{@{}lcccc@{}}
\toprule
 (MER)            & PASS  & WIN  & Average & Rank \\ \midrule
Claude2     & 100.0 & 81.3 & 90.6    & 1    \\
BLOOM-176B   & 97.5  & 65.4 & 81.4    & 2    \\
LLaMA2-70B  & 95.0  & 64.5 & 79.7    & 3    \\
Vicuna-33B   & 92.5  & 50.0 & 71.3    & 4    \\
GPT4         & 90.0  & 50.0 & 70.0    & 5    \\
GPT3.5       & 92.5  & 28.4 & 60.4    & 6    \\
Falcon-180B & 91.3  & 19.2 & 55.2    & 7    \\
Vicuna-7B    & 73.8  & 15.3 & 44.5    & 8    \\
LLaMA2-7B   & 67.5  & 14.8 & 41.2    & 9    \\
Falcon-7B    & 57.5  & 13.0 & 35.3    & 10   \\
BLOOM-7B     & 2.5   & 0.0  & 1.3     & 11   \\ \bottomrule
\end{tabular}%
}
\caption{Performance of different LLMs in mixed event recognition (MER).}
\label{tab:qingshang-exp1-2}
\end{table}

\begin{figure}[!h]
  \centering
  \includegraphics[width=\linewidth]{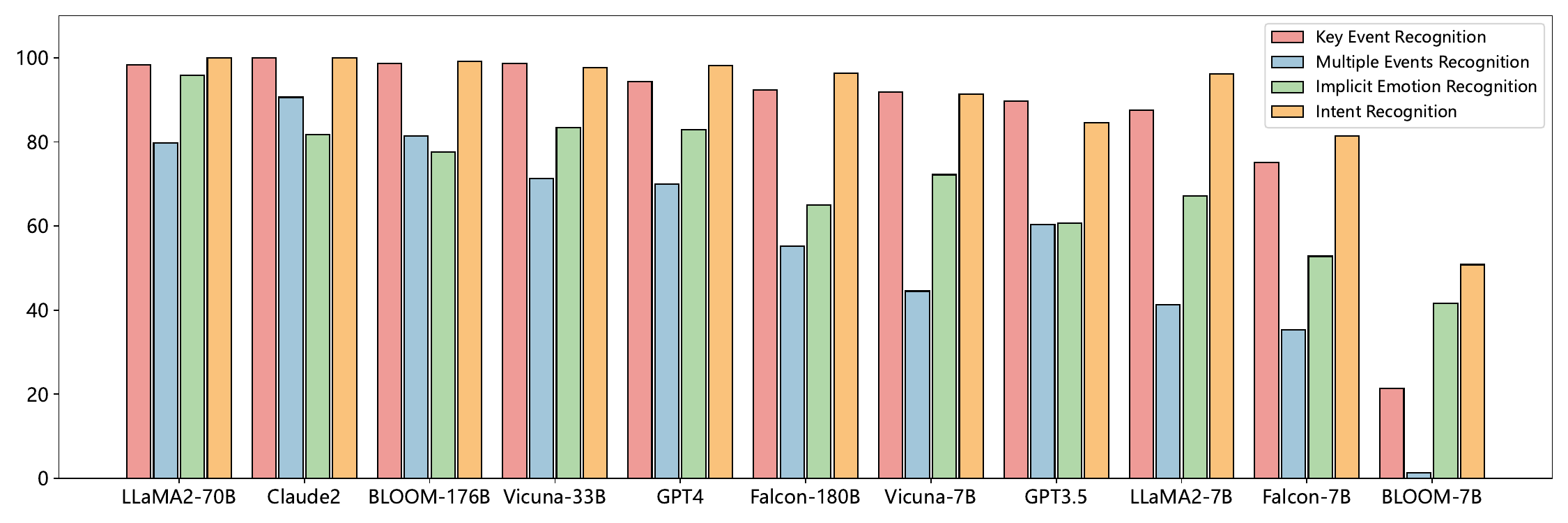}
  \caption{The overall performance of different LLMs in the proposed EmotionQueen benchmark.}
  \label{fig:qingshang-exp1}
\end{figure}

Performance of different LLMs across four tasks is shown in Table~\ref{tab:qingshang-exp1-1}, Table~\ref{tab:qingshang-exp1-2}, Table~\ref{tab:qingshang-exp1-3} and Table~\ref{tab:qingshang-exp1-4}. We rank them based on the average of PASS rate and WIN rate, and compare the their comprehensive empathy capabilities as shown in Table~\ref{tab:qingshang-exp1} and Fig.~\ref{fig:qingshang-exp1}.
In the aspect of key event recognition (denoted as ``KER''), Claude2, BLOOM-176B, and LLaMA2-70B show nearly perfect PASS rate, demonstrating their strong capability in capturing the core events of user statements. In WIN rate, Claude2 maintains a lead, other LLMs such as Vicuna-33b and LLaMA2-70B fall slightly short. 
% This suggests that there is room for improvement in their response methods.
In the aspect of mixed event recognition (denoted as ``MER''), Claude2 also excels in identifying multiple events, showing its capability in handling complex scenarios. But the WIN rate generally decreases for all LLMs, suggesting potential room for improvement in understanding and balanced responses to mixed events.
Regarding implicit emotion recognition (denoted as ``IER''), LLaMA2-70B, Vicuna-33B, and GPT4 perform well. However, the WIN rate analysis shows that, except for LLaMA2-70B, other LLMs like Vicuna-33B and GPT4 have gaps in response appropriateness.
% , highlighting the need for improvement in accurately responding to user emotions.
In intent recognition (denoted as ``IR''), most LLMs, especially Claude2 and LLaMA2-70B, show a high PASS rate, but they still show potential in providing effective guidance.
Overall, although most LLMs excel in identifying the key content of user statements, they still have room for improvement in providing empathetic responses. 
% This analysis reveals key directions for future development of AI models, especially in enhancing the win rate and appropriateness of responses.
% Similarly, we find that larger-sized models of the same type tend to perform better.

% Please add the following required packages to your document preamble:
% \usepackage{booktabs}
% \usepackage{graphicx}
\begin{table}[]
\centering
\resizebox{0.3\textwidth}{!}{%
\begin{tabular}{@{}lcccc@{}}
\toprule
  (IER)           & PASS  & WIN  & Average & Rank \\ \midrule
LLaMA2-70B  & 99.1  & 92.5 & 95.8    & 1    \\
Vicuna-33B   & 99.1  & 67.7 & 83.4    & 2    \\
GPT4         & 99.1  & 66.7 & 82.9    & 3    \\
Claude2     & 100.0 & 63.6 & 81.8    & 4    \\
BLOOM-176B   & 100.0 & 55.1 & 77.6    & 5    \\
Vicuna-7B    & 97.2  & 47.1 & 72.2    & 6    \\
LLaMA2-7B   & 96.3  & 37.9 & 67.1    & 7    \\
Falcon-180B & 100.0 & 29.9 & 65.0    & 8    \\
GPT3.5       & 100.0 & 21.5 & 60.7    & 9    \\
Falcon-7B    & 83.2  & 22.5 & 52.8    & 10   \\
BLOOM-7B     & 78.5  & 4.8  & 41.6    & 11   \\ \bottomrule
\end{tabular}%
}
\caption{Performance of different LLMs in implicit emotion recognition (IER).}
\label{tab:qingshang-exp1-3}
\end{table}

% Please add the following required packages to your document preamble:
% \usepackage{booktabs}
% \usepackage{graphicx}
\begin{table}[]
\centering
\resizebox{0.3\textwidth}{!}{%
\begin{tabular}{@{}lcccc@{}}
\toprule
  (IR)           & PASS  & WIN   & Average & Rank \\ \midrule
Claude2     & 100.0 & 100.0 & 100.0   & 1    \\
LLaMA2-70B  & 100.0 & 99.7  & 99.9    & 2    \\
BLOOM-176B   & 100.0 & 98.1  & 99.1    & 3    \\
GPT4         & 100.0 & 96.3  & 98.1    & 4    \\
Vicuna-33B   & 100.0 & 95.3  & 97.7    & 5    \\
Falcon-180B & 100.0 & 92.5  & 96.3    & 6    \\
LLaMA2-7B   & 97.3  & 95.1  & 96.2    & 7    \\
Vicuna-7B    & 96.7  & 86.0  & 91.3    & 8    \\
GPT3.5       & 100.0 & 69.2  & 84.6    & 9    \\
Falcon-7B    & 98.1  & 64.8  & 81.4    & 10   \\
BLOOM-7B     & 75.7  & 25.9  & 50.8    & 11   \\ \bottomrule
\end{tabular}%
}
\caption{Performance of different LLMs in intent recognition (IR).}
\label{tab:qingshang-exp1-4}
\end{table}

% Please add the following required packages to your document preamble:
% \usepackage{booktabs}
% \usepackage{graphicx}
\begin{table}[]
\centering
\resizebox{0.3\textwidth}{!}{%
\begin{tabular}{@{}lcccl@{}}
\toprule
 (Total)           & PASS  & WIN  & Average & Rank \\ \midrule
LLaMA2-70B  & 98.5  & 88.3 & 93.4    & 1    \\
Claude2     & 100.0 & 86.2 & 93.1    & 2    \\
BLOOM-176B  & 99.4  & 79.0 & 89.2    & 3    \\
Vicuna-33B  & 97.7  & 77.8 & 87.8    & 4    \\
GPT4        & 97.1  & 75.7 & 86.4    & 5    \\
Falcon-180B & 97.6  & 56.8 & 77.2    & 6    \\
Vicuna-7B   & 91.5  & 58.5 & 75.0    & 7    \\
GPT3.5      & 97.7  & 50.0 & 73.9    & 8    \\
LLaMA2-7B   & 89.8  & 56.1 & 73.0    & 9    \\
Falcon-7B   & 81.3  & 41.0 & 61.2    & 10   \\
Bloom-7B    & 48.7  & 8.8  & 28.7    & 11   \\ \bottomrule
\end{tabular}%
}
\caption{The overall performance of different LLMs in the proposed EmotionQueen benchmark.}
\label{tab:qingshang-exp1}
\end{table}

We also explore the emotional intelligence of these LLMs on the traditional emotion datasets, including MELD~\citep{poria2018meld}, EmotionLines~\citep{chen2018emotionlines}, and DailyDialog~\citep{li2017dailydialog}, which are all about emotion categorization. We randomly select 100 samples from each for assessment with accuracy as the metric. We find that the results as shown in Table~\ref{tab:qingshang-sup1-1} resemble the rankings in Implicit Emotion Recognition's PASS, suggesting that our tasks, inclusive of these emotion categorization tasks, mirror the challenge of recognizing implicit emotions. This confirms that our task setup effectively encompasses these types of emotional recognition challenges.

\begin{table}[H]
\centering
\resizebox{0.44\textwidth}{!}{%
\begin{tabular}{@{}lccccc@{}}
\toprule
LLM         & MELD  & Emotionlines & DailyDialog & Average & Rank \\ \midrule
Claude2     & 85.23 & 83.67        & 84.26       & 84.39   & 1    \\
BLOOM-176B  & 84.89 & 82.77        & 83.59       & 83.75   & 2    \\
GPT3.5      & 84.98 & 83.55        & 82.16       & 83.56   & 3    \\
GPT4        & 84.36 & 81.37        & 80.55       & 82.09   & 4    \\
LLaMA2-70B  & 81.55 & 78.76        & 80.12       & 80.14   & 5    \\
Falcon-180B & 79.47 & 80.21        & 79.44       & 79.71   & 6    \\
Vicuna-33B  & 77.44 & 79.34        & 76.58       & 77.79   & 7    \\
LLaMA2-7B   & 73.66 & 75.36        & 72.01       & 73.68   & 8    \\
Vicuna-7B   & 75.23 & 73.46        & 71.31       & 73.33   & 9    \\
Falcon-7B   & 71.54 & 70.34        & 68.44       & 70.11   & 10   \\
BLOOM-7B    & 69.37 & 67.59        & 70.45       & 69.14   & 11   \\ \bottomrule
\end{tabular}%
}
\caption{Performance of each LLMs in other public datasets.}
\label{tab:qingshang-sup1-1}
\end{table}

%%%%%779P问题1结束，放在这里主要是对比table5

Moreover, in-context learning (ICL) which introduces a human-written sample into the prompt~\citep{chen2023mapo} typically improves LLMs' performance as shown in Table~\ref{tab:qingshang-exp2}, while most LLMs show a decline in the 0-shot setting, demonstrating its critical role in enhancing the quality of recognition and response.

\begin{table}[]
\centering
\resizebox{0.38\textwidth}{!}{%
\begin{tabular}{@{}lccc@{}}
\toprule
             & Average (ICL) & Average (0-shot) & ↓(\%) \\ \midrule
BLOOM-7B     & 28.7          & 26.0             & 10.8  \\
Falcon-7B    & 61.2          & 58.4             & 4.8   \\
LLaMA2-7B   & 73.0          & 69.7             & 4.7   \\
Falcon-180B & 77.2          & 74.2             & 4.1   \\
Vicuna-7B    & 75.0          & 72.8             & 3.0   \\
GPT3.5       & 73.9          & 71.9             & 2.8   \\
GPT4         & 86.4          & 84.3             & 2.4   \\
BLOOM-176B   & 89.2          & 87.4             & 2.1   \\
Vicuna-33B   & 87.8          & 86.1             & 2.0   \\
LLaMA2-70B  & 93.4          & 92.0             & 1.6   \\
Claude2     & 93.1          & 92.2             & 1.0   \\ \bottomrule
\end{tabular}%
}
\caption{The comparison between the performance of in-context learning and zero-shot settings.}
\label{tab:qingshang-exp2}
\end{table}

\emph{Question 2: What's the relationship between PASS and WIN rate? Answer 2: There is almost no correlation!}

\begin{figure*}[!h]
  \centering
  \includegraphics[width=0.8\linewidth]{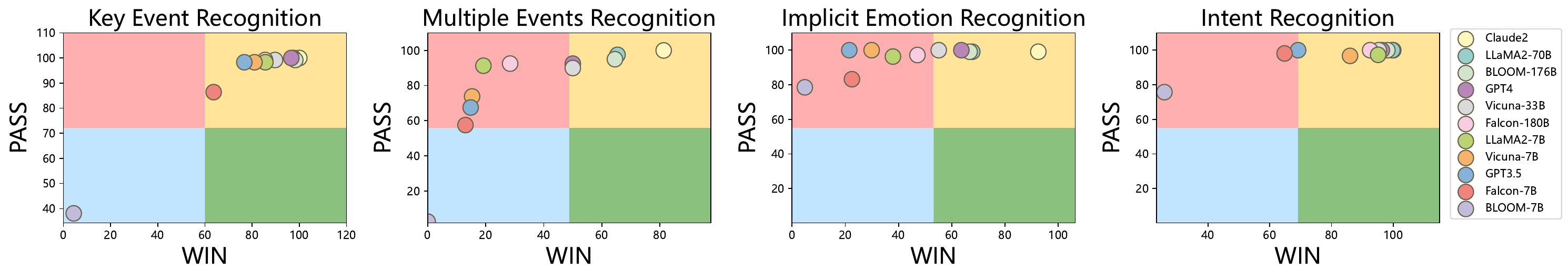}
  \caption{The relationship between PASS rate and WIN rate of different LLMs in four tasks, respectively.}
  \label{fig:qingshang-exp2}
\end{figure*}

We also analyze the relationship between PASS and WIN rate in four tasks as shown in Fig.~\ref{fig:qingshang-exp2}. 
It suggests that there is little correlation between the PASS and WIN rate across different LLMs for the various tasks. In the Key Event Recognition task, most LLMs cluster around high PASS scores and high WIN scores, indicating their strong capabilities in this task. 
In the Mixed Events Recognition task, when observing a low WIN rate (below 30\%), there appears to be a positive correlation between PASS and WIN rate, with a wide dispersion in WIN rates among the LLMs. However, Claude2 still stands out, achieving high PASS rate and WIN rate compared to others.
In Implicit Emotion Recognition and Intent Recognition tasks, there's virtually no correlation between PASS and WIN rates, with PASS rates being generally high across the board. For Implicit Emotion Recognition, WIN rates are distributed across various scores, whereas for Intent Recognition, there is a concentration of higher WIN rates.
Overall, LLaMA-70B and Claude2 show consistent and superior performance across all tasks and metrics, while other models have their strengths in specific areas. 

\emph{Question 3: Which LLM wins in each domain? Answer 3: LLaMA-70B in economic status and accidents, as well as Claude2 in achievements, family and friends, and health status!}

\begin{figure*}[!h]
  \centering
  \includegraphics[width=0.9\linewidth]{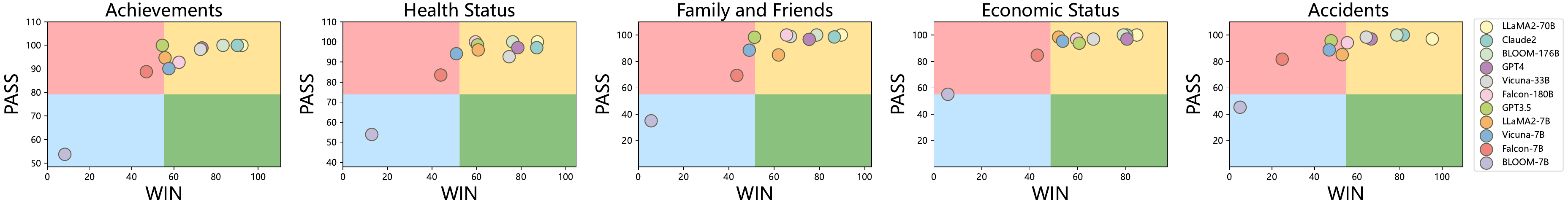}
  \caption{The relationship between PASS rate and WIN rate of different LLMs in five categories of events, respectively.}
  \label{fig:qingshang-exp3}
\end{figure*}

We also compare the performance of different models in different categories of events as shown in Table~\ref{tab:qingshang-exp3-1}, Table~\ref{tab:qingshang-exp3-2}, Table~\ref{tab:qingshang-exp3-3} and Table~\ref{tab:qingshang-exp3-4}. We also rank them based on the average of PASS rate and WIN rate.
In the category of the achievement (denoted as ``Ach''), Claude2, LLaMA2-70B, and BLOOM-176B all show a 100\% pass rate, but Claude2 slightly leads in win rate.
In the category of the health status (denoted as ``H''), Claude2 continues to hold the highest average score, with LLaMA2-70B and BLOOM-176B also performing well. 
In the category of the family and friends (denoted as ``F''), Claude2 and LLaMA2-70B continue to lead, showing their ability to handle complex emotional issues related to close relationships. 
In the category of the economic status (denoted as ``E''), LLaMA2-70B achieves the highest score. Claude2 and GPT4 also show strong performance, particularly in PASS rate.
Finally, in the category of accidents (denoted as ``Acc''), LLaMA2-70B leads significantly in WIN rate, indicating its ability to provide more accurate and effective responses in sudden and bad situations. 
We also shown the relationship between PASS rate and WIN rate in each scenario of event in Fig.~\ref{fig:qingshang-exp3}. The trend is similar as that overall relationship as shown in Fig.~\ref{fig:qingshang-exp2}

% Please add the following required packages to your document preamble:
% \usepackage{booktabs}
% \usepackage{graphicx}
\begin{table}[]
\centering
\resizebox{0.3\textwidth}{!}{%
\begin{tabular}{@{}lcccc@{}}
\toprule
 (Ach)            & PASS  & WIN  & Average & Rank \\ \midrule
Claude2     & 100.0 & 92.3 & 96.1    & 1    \\
LLaMA2-70B  & 100.0 & 90.1 & 95.1    & 2    \\
BLOOM-176B   & 100.0 & 83.3 & 91.7    & 3    \\
GPT4         & 98.9  & 73.3 & 86.1    & 4    \\
Vicuna-33B   & 98.3  & 72.6 & 85.5    & 5    \\
LLaMA2-7B   & 92.8  & 62.4 & 77.6    & 6    \\
Falcon-180B & 100.0 & 54.5 & 77.3    & 7    \\
GPT3.5       & 94.7  & 55.8 & 75.2    & 8    \\
Vicuna-7B    & 90.1  & 57.6 & 73.8    & 9    \\
Falcon-7B    & 88.8  & 46.9 & 67.9    & 10   \\
BLOOM-7B     & 53.7  & 8.2  & 30.9    & 11   \\ \bottomrule
\end{tabular}%
}
\caption{The performance of models in the first category of the achievement.}
\label{tab:qingshang-exp3-1}
\end{table}

% Please add the following required packages to your document preamble:
% \usepackage{booktabs}
% \usepackage{graphicx}
\begin{table}[]
\centering
\resizebox{0.3\textwidth}{!}{%
\begin{tabular}{@{}lcccc@{}}
\toprule
   (H)           & PASS  & WIN  & Average & Rank \\ \midrule
Claude2     & 100.0 & 87.3 & 93.6    & 1    \\
LLaMA2-70B  & 97.1  & 87.0 & 92.0    & 2    \\
BLOOM-176B   & 100.0 & 76.2 & 88.1    & 3    \\
Vicuna-33B   & 97.1  & 78.5 & 87.8    & 4    \\
GPT4         & 92.6  & 74.6 & 83.6    & 5    \\
GPT3.5       & 100.0 & 59.5 & 79.8    & 6    \\
Falcon-180B & 98.5  & 60.4 & 79.4    & 7    \\
Vicuna-7B    & 96.0  & 60.7 & 78.4    & 8    \\
LLaMA2-7B   & 94.1  & 50.9 & 72.5    & 9    \\
Falcon-7B    & 83.5  & 43.9 & 63.7    & 10   \\
BLOOM-7B     & 53.9  & 12.9 & 33.4    & 11   \\ \bottomrule
\end{tabular}%
}
\caption{Performance of different LLMs in the task related to health status (H).}
\label{tab:qingshang-exp3-2}
\end{table}

% Please add the following required packages to your document preamble:
% \usepackage{booktabs}
% \usepackage{graphicx}
\begin{table}[]
\centering
\resizebox{0.3\textwidth}{!}{%
\begin{tabular}{@{}lcccc@{}}
\toprule
   (F)           & PASS  & WIN  & Average & Rank \\ \midrule
Claude2     & 100.0 & 89.7 & 94.9    & 1    \\
LLaMA2-70B  & 98.5  & 86.6 & 92.6    & 2    \\
BLOOM-176B   & 100.0 & 78.8 & 89.4    & 3    \\
GPT4         & 96.7  & 75.5 & 86.1    & 4    \\
Vicuna-33B   & 98.9  & 67.2 & 83.1    & 5    \\
Falcon-180B & 100.0 & 65.5 & 82.7    & 6    \\
GPT3.5       & 98.3  & 51.3 & 74.8    & 7    \\
LLaMA2-7B   & 84.9  & 61.9 & 73.4    & 8    \\
Vicuna-7B    & 88.5  & 49.0 & 68.8    & 9    \\
Falcon-7B    & 69.4  & 43.5 & 56.5    & 10   \\
BLOOM-7B     & 35.0  & 5.6  & 20.3    & 11   \\ \bottomrule
\end{tabular}%
}
\caption{Performance of different LLMs in the task related to family and friends (F). }
\label{tab:qingshang-exp3-3}
\end{table}

\emph{Question 4: Is the automatic scores generated by GPT4 agree with human perspectives? Answer 4: Yes, and the correlation coefficient reaches 0.991!}

\begin{figure}[!t]
  \centering
  \includegraphics[width=0.9\linewidth]{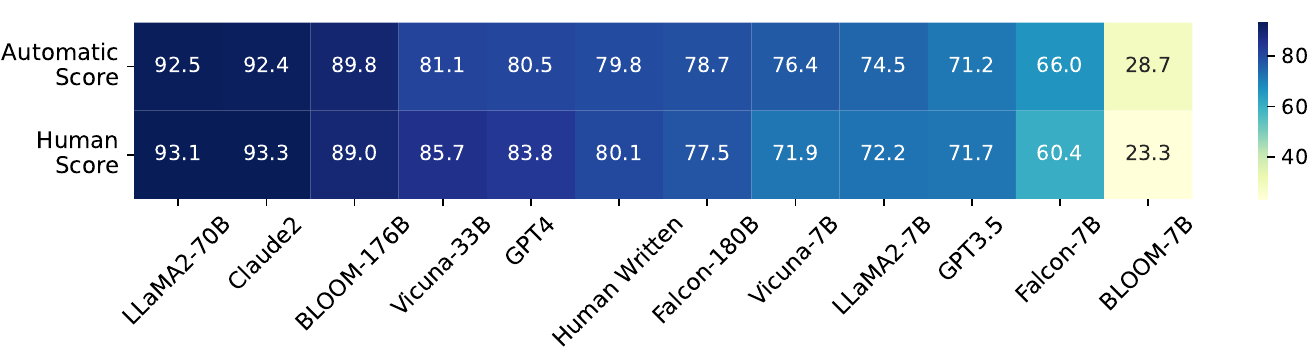}
  \caption{The Pearson correlation coefficient that normalized to a 1-100 scale between automatic scores and human scores of different LLMs.}
  \label{fig:qingshang-exp4}
\end{figure}

We select 1000 statements with an equal number in each task to make human evaluation. We compare the consistency between automatic scores and human ratings as shown Fig.~\ref{fig:qingshang-exp4}. The Pearson correlation coefficient achieves 0.991, indicating a strong alignment. 
We also recruit three volunteers to respond to these statements, selecting the one with the highest average score for WIN and PASS as the human-written performance. We find that some LLMs, such as LLaMA2-70B and Claude2, have already surpassed the real human in
empathy capabilities, at least in the tasks in EmotionQueen.
This indicates that some LLMs are capable of generating responses that can exceed the emotional value brought by human interactions.

% Please add the following required packages to your document preamble:
% \usepackage{booktabs}
% \usepackage{graphicx}
\begin{table}[]
\centering
\resizebox{0.3\textwidth}{!}{%
\begin{tabular}{@{}lcccc@{}}
\toprule
   (E)           & PASS  & WIN  & Average & Rank \\ \midrule
LLaMA2-70B  & 100.0 & 84.6 & 92.3    & 1    \\
Claude2     & 100.0 & 80.4 & 90.2    & 2    \\
GPT4         & 100.0 & 79.0 & 89.5    & 3    \\
BLOOM-176B   & 96.9  & 80.5 & 88.7    & 4    \\
Vicuna-33B   & 96.9  & 66.5 & 81.7    & 5    \\
Falcon-180B & 96.9  & 59.5 & 78.2    & 6    \\
LLaMA2-7B   & 93.8  & 60.6 & 77.2    & 7    \\
GPT3.5       & 98.4  & 52.0 & 75.2    & 8    \\
Vicuna-7B    & 95.3  & 53.7 & 74.5    & 9    \\
Falcon-7B    & 84.7  & 43.1 & 63.9    & 10   \\
BLOOM-7B     & 55.1  & 5.8  & 30.5    & 11   \\ \bottomrule
\end{tabular}%
}
\caption{Performance of different LLMs in the task related to economic status (E).}
\label{tab:qingshang-exp3-4}
\end{table}

% Please add the following required packages to your document preamble:
% \usepackage{booktabs}
% \usepackage{graphicx}
\begin{table}[]
\centering
\resizebox{0.3\textwidth}{!}{%
\begin{tabular}{@{}lcccc@{}}
\toprule
   (Acc)           & PASS  & WIN  & Average & Rank \\ \midrule
LLaMA2-70B  & 97.0  & 95.3 & 96.1    & 1    \\
Claude2     & 100.0 & 81.7 & 90.9    & 2    \\
BLOOM-176B   & 100.0 & 78.7 & 89.3    & 3    \\
GPT4         & 97.0  & 66.6 & 81.8    & 4    \\
Vicuna-33B   & 98.4  & 64.3 & 81.4    & 5    \\
Falcon-180B & 94.1  & 55.4 & 74.8    & 6    \\
GPT3.5       & 95.6  & 47.8 & 71.7    & 7    \\
LLaMA2-7B   & 85.1  & 53.0 & 69.0    & 8    \\
Vicuna-7B    & 88.7  & 46.9 & 67.8    & 9    \\
Falcon-7B    & 81.7  & 24.8 & 53.2    & 10   \\
BLOOM-7B     & 45.3  & 5.0  & 25.2    & 11   \\ \bottomrule
\end{tabular}%
}
\caption{Performance of different LLMs in the task related to accidents (Acc).}
\label{tab:qingshang-exp3-5}
\end{table}

\subsection{Case Study}

\begin{figure*}[!h]
  \centering
  \includegraphics[width=0.77\linewidth]{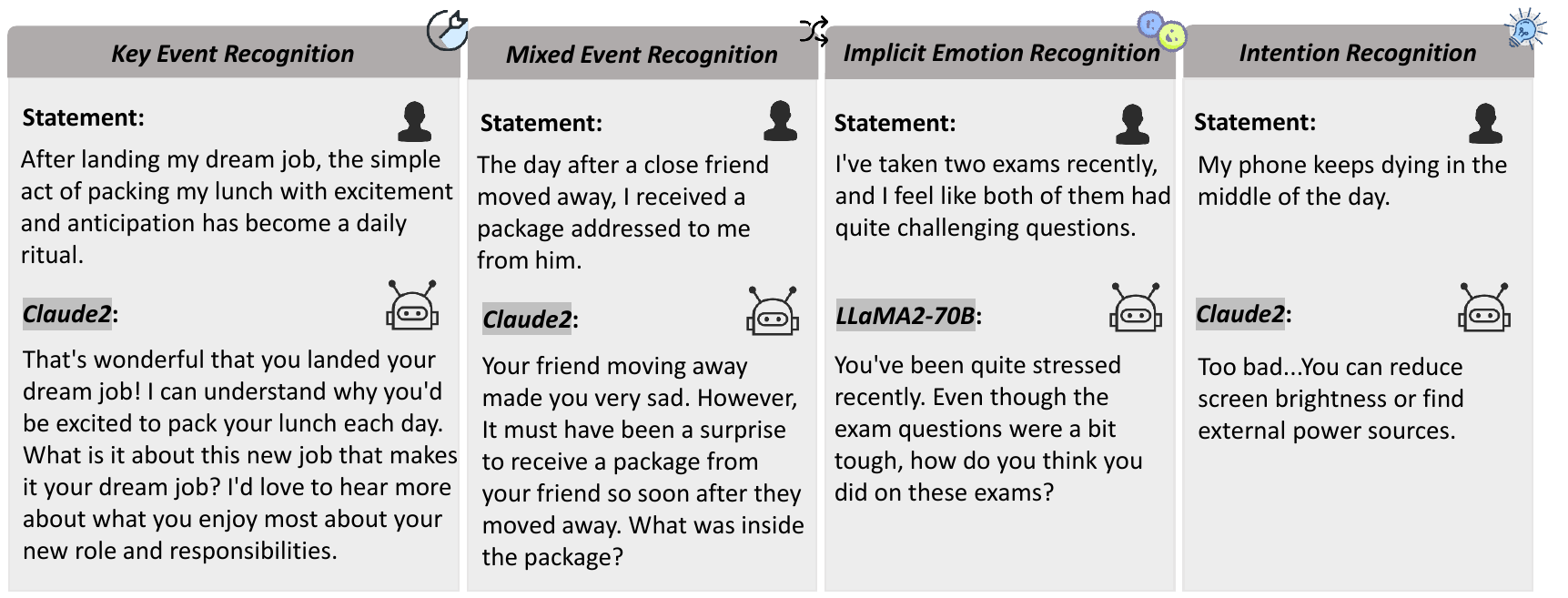}
  \caption{Responses of top LLMs, i.e. Claude2 and LLaMA2-70B, in four tasks for given statements.}
  \label{fig:qingshang-case}
\end{figure*}

\begin{figure*}[!h]
  \centering
  \includegraphics[width=0.8\linewidth]{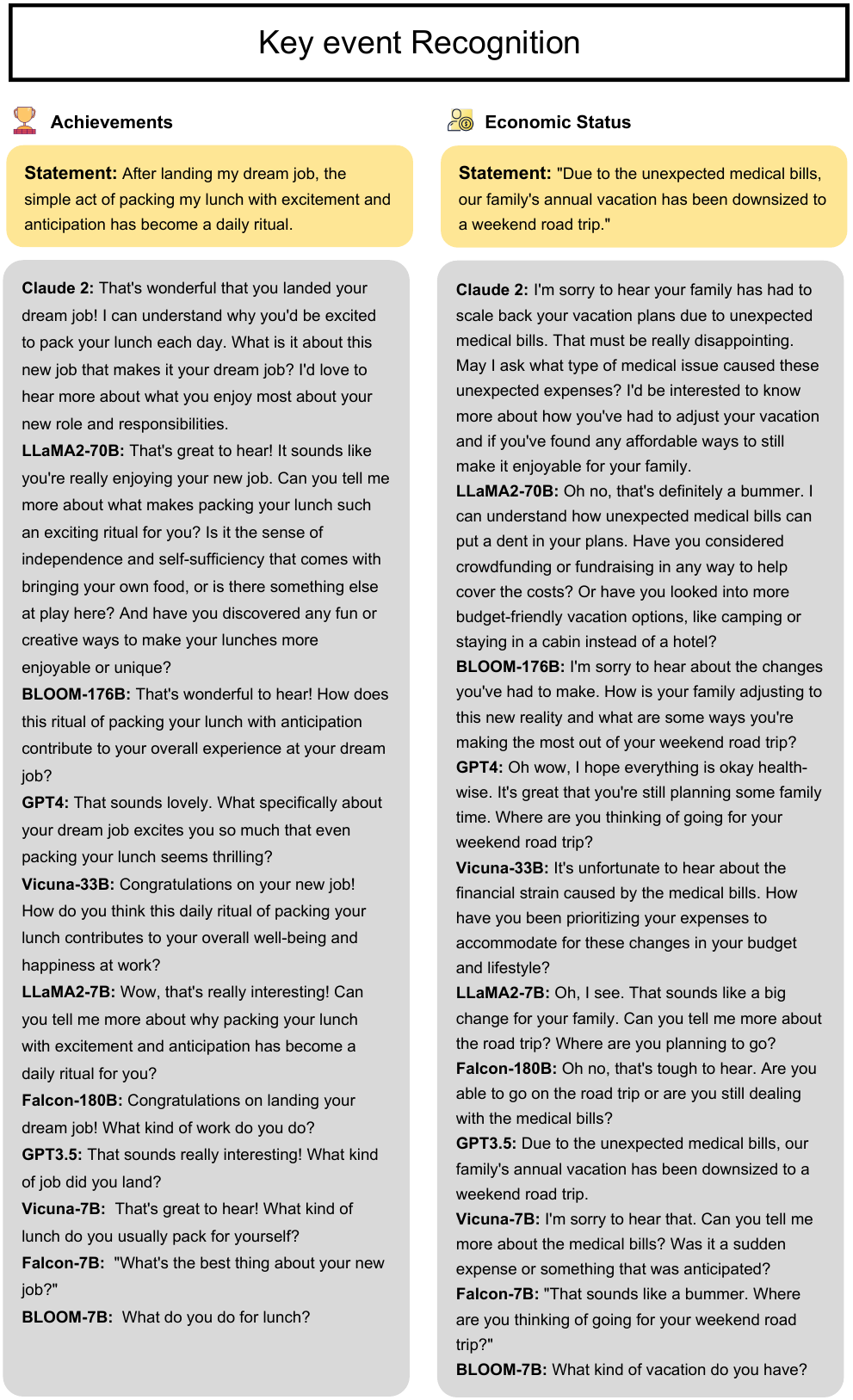}
  \caption{Responses of different LLMs in key event recognition for given statements.}
  \label{fig:1qingshang-case1}
\end{figure*}

\begin{figure*}[!h]
  \centering
  \includegraphics[width=0.8\linewidth]{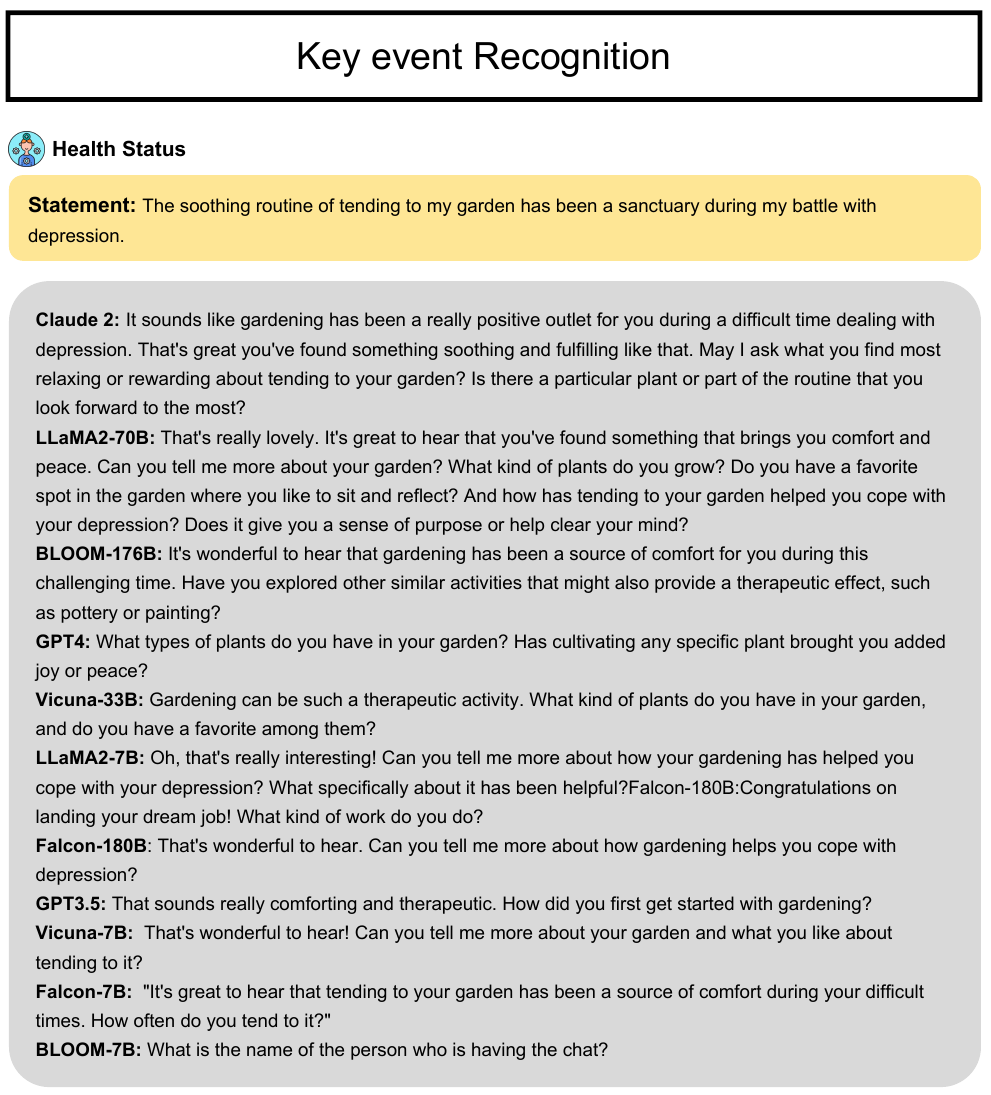}
  \caption{Responses of different LLMs in key event recognition for given statements.}
  \label{fig:1qingshang-case2}
\end{figure*}

\begin{figure*}[!h]
  \centering
  \includegraphics[width=0.8\linewidth]{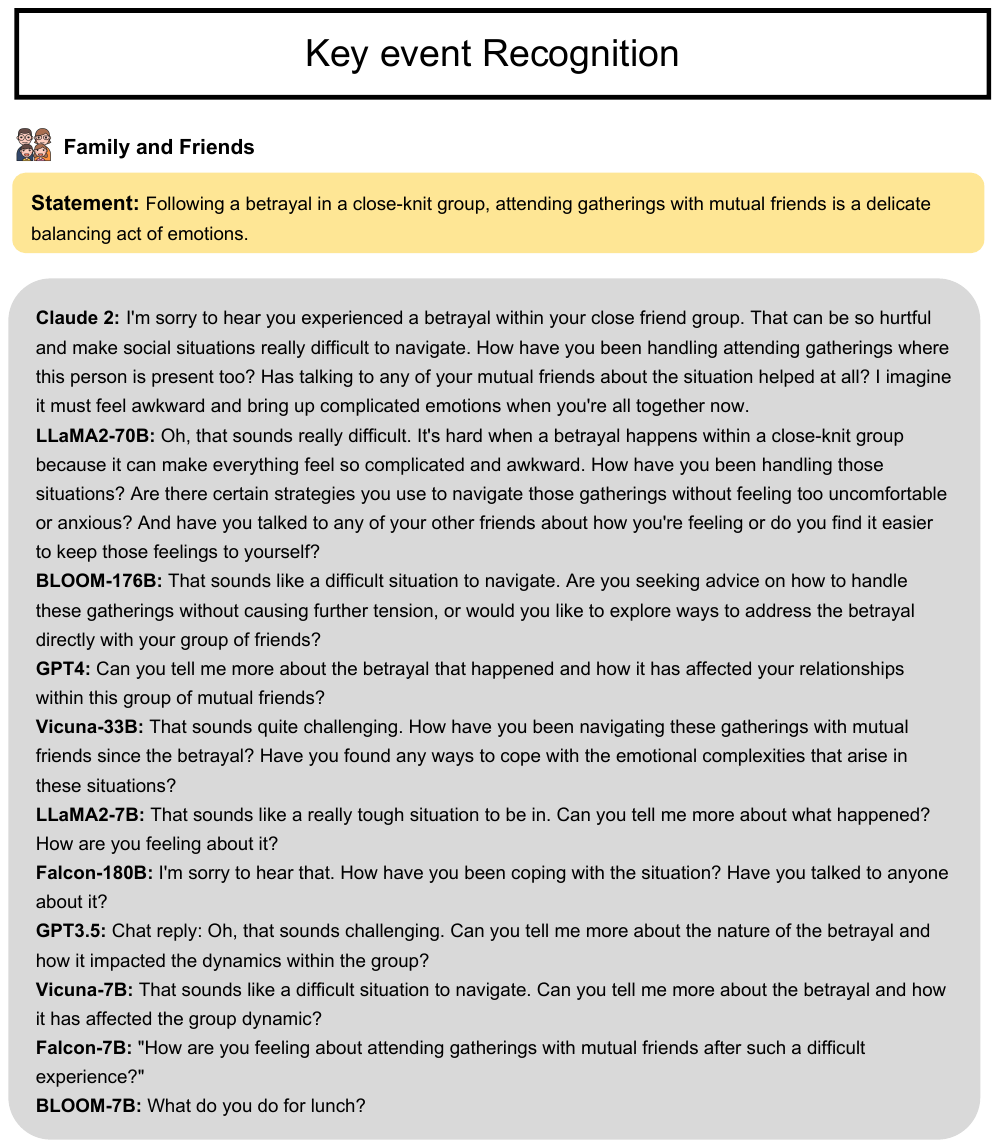}
  \caption{Responses of different LLMs in key event recognition for given statements.}
  \label{fig:1qingshang-case3}
\end{figure*}

\begin{figure*}[!h]
  \centering
  \includegraphics[width=0.8\linewidth]{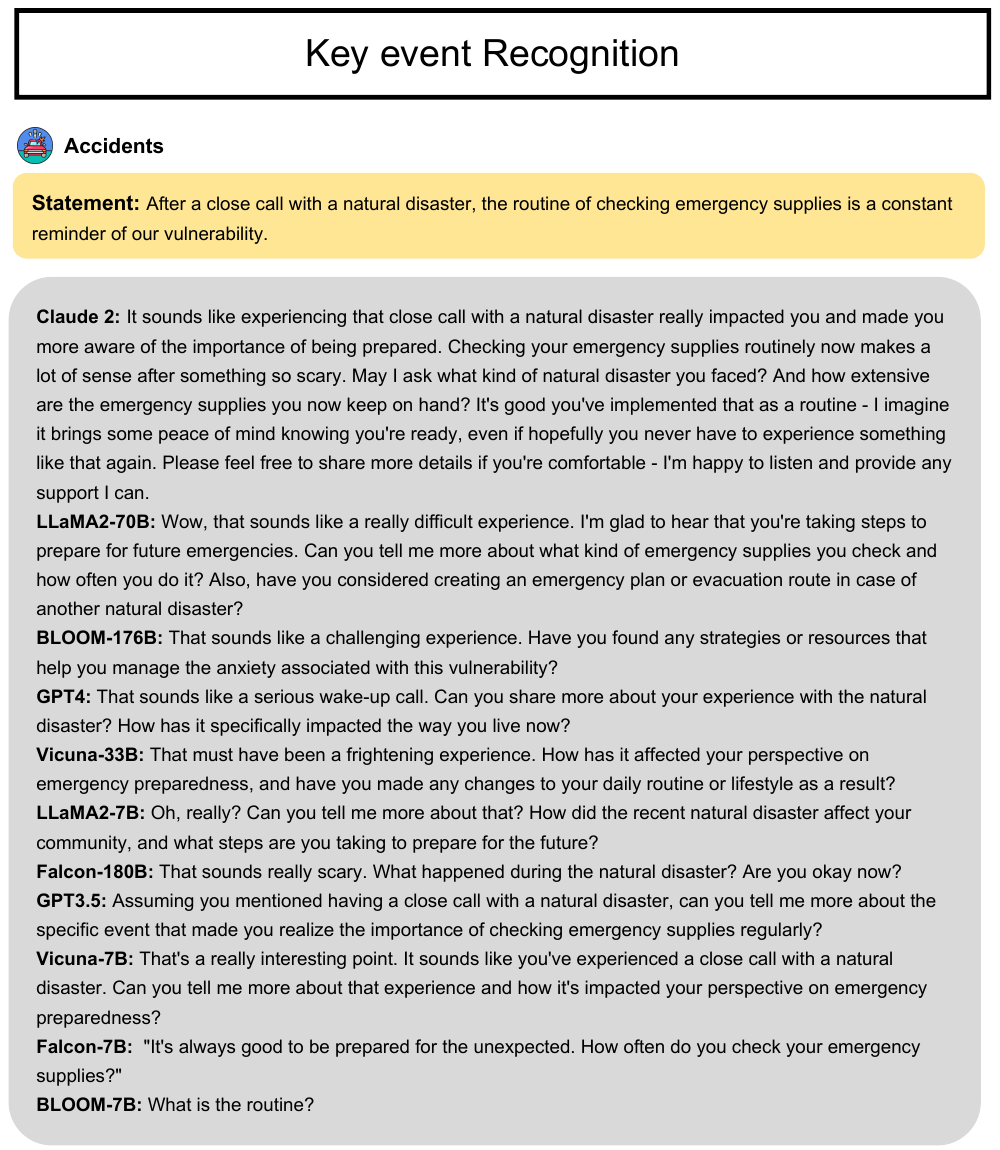}
  \caption{Responses of different LLMs in key event recognition for given statements.}
  \label{fig:1qingshang-case4}
\end{figure*}

\begin{figure*}[!h]
  \centering
  \includegraphics[width=0.8\linewidth]{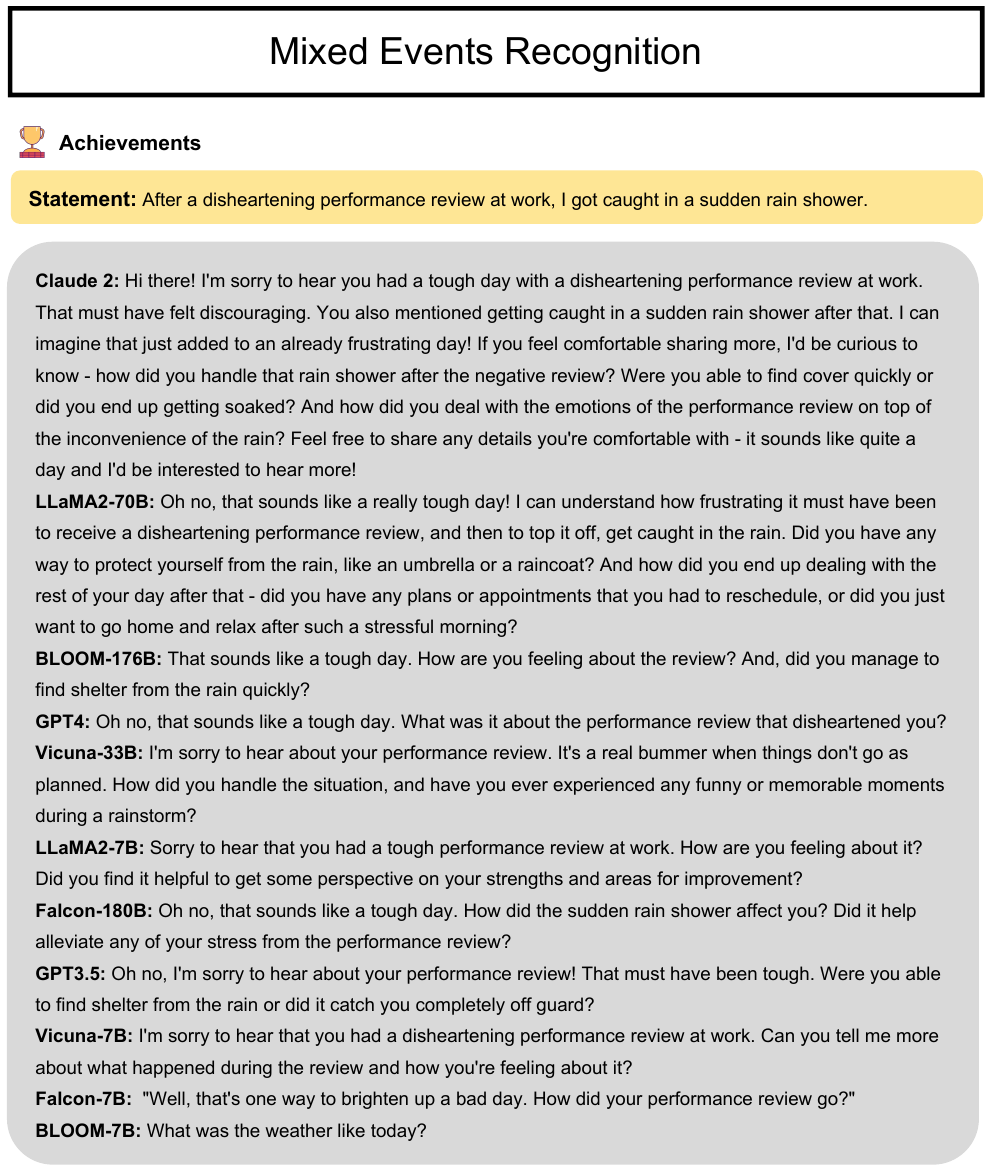}
  \caption{Responses of different LLMs in mixed events recognition for given statements.}
  \label{fig:2qingshang-case1}
\end{figure*}

\begin{figure*}[!h]
  \centering
  \includegraphics[width=0.8\linewidth]{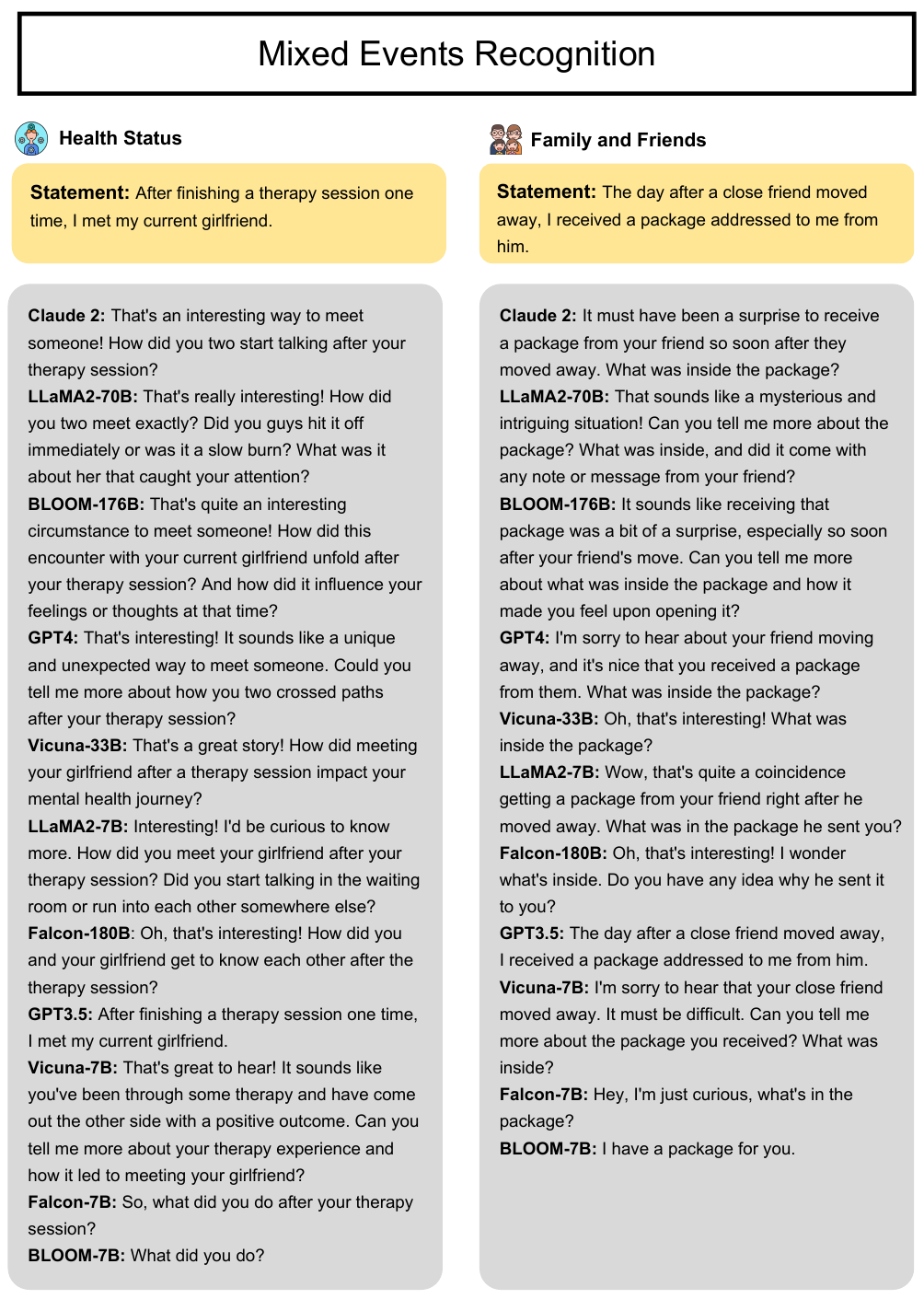}
  \caption{Responses of different LLMs in mixed events recognition for given statements.}
  \label{fig:2qingshang-case2}
\end{figure*}

\begin{figure*}[!h]
  \centering
  \includegraphics[width=0.8\linewidth]{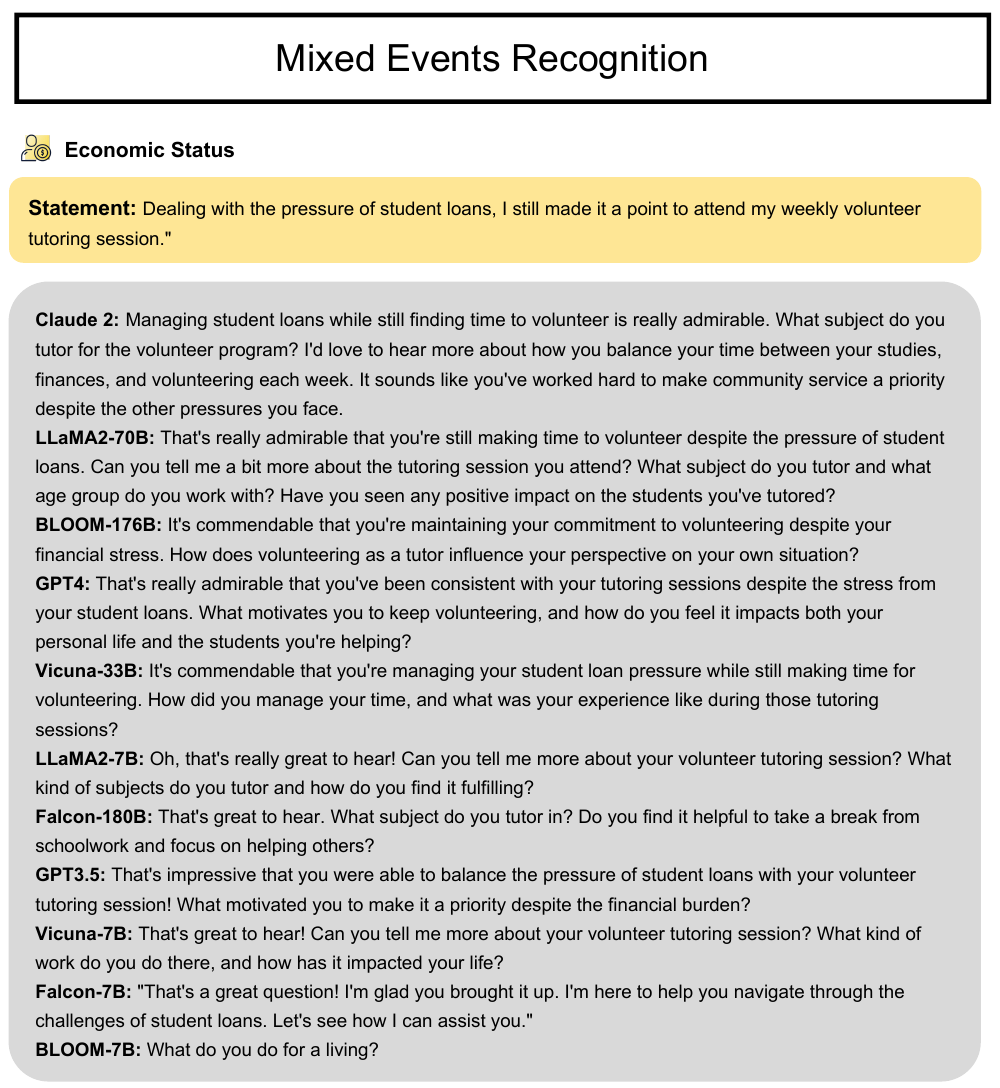}
  \caption{Responses of different LLMs in mixed events recognition for given statements.}
  \label{fig:2qingshang-case3}
\end{figure*}

\begin{figure*}[!h]
  \centering
  \includegraphics[width=0.8\linewidth]{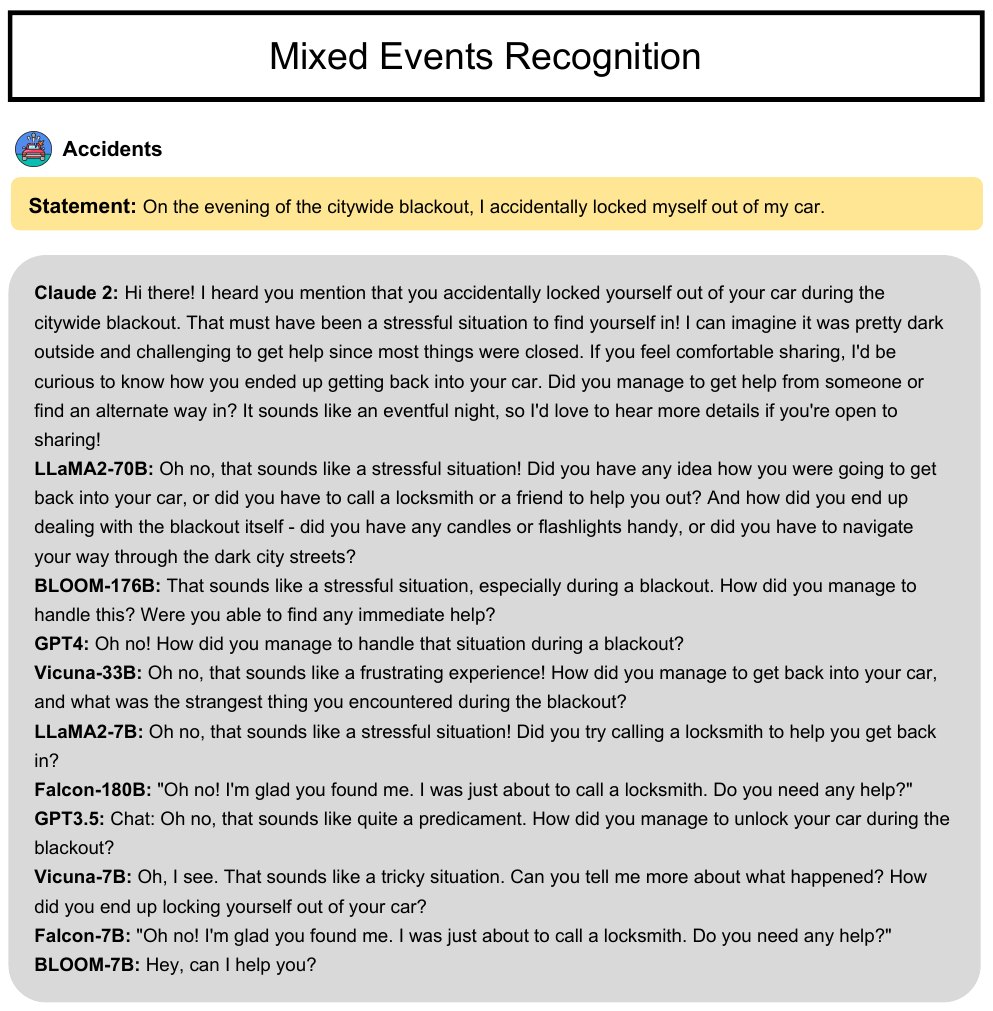}
  \caption{Responses of different LLMs in mixed events recognition for given statements.}
  \label{fig:2qingshang-case4}
\end{figure*}

\begin{figure*}[!h]
  \centering
  \includegraphics[width=0.8\linewidth]{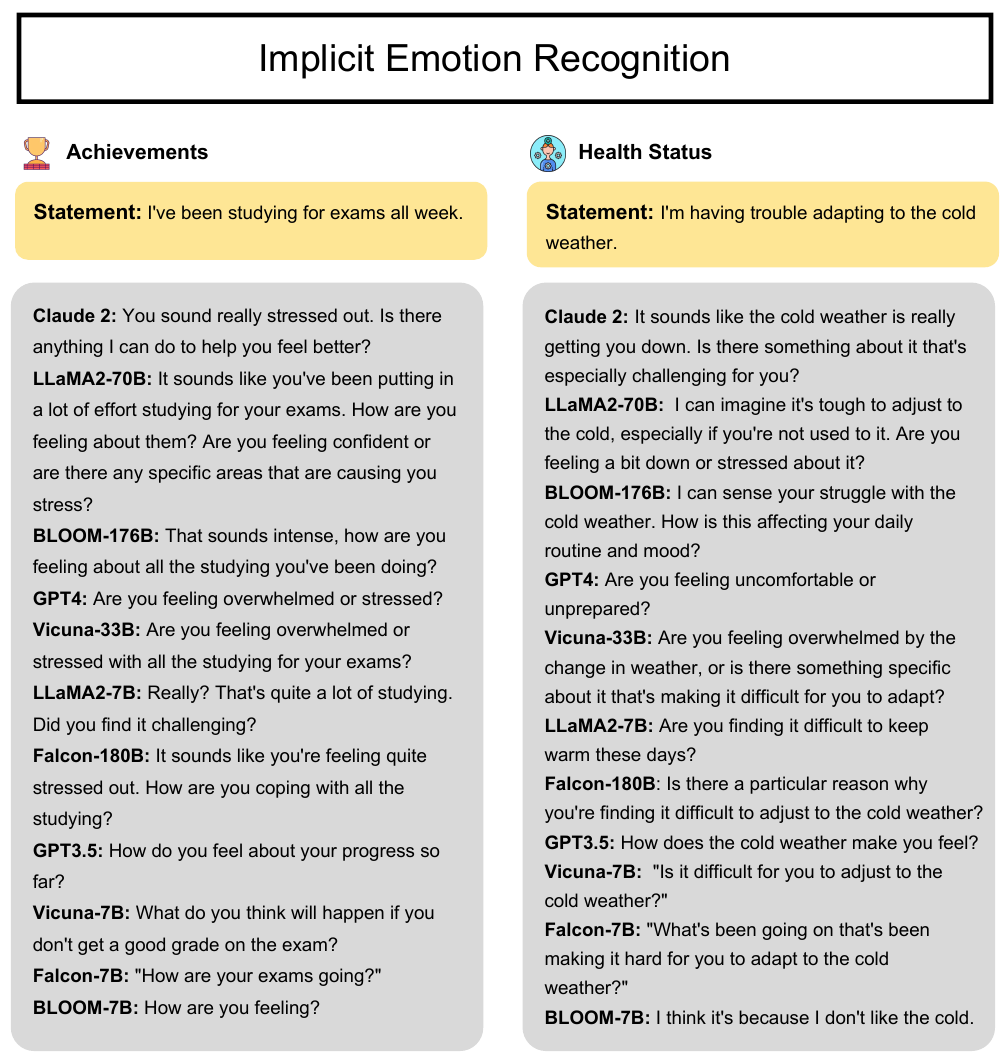}
  \caption{Responses of different LLMs in implicit emotion recognition for given statements.}
  \label{fig:3qingshang-case1}
\end{figure*}

\begin{figure*}[!h]
  \centering
  \includegraphics[width=0.8\linewidth]{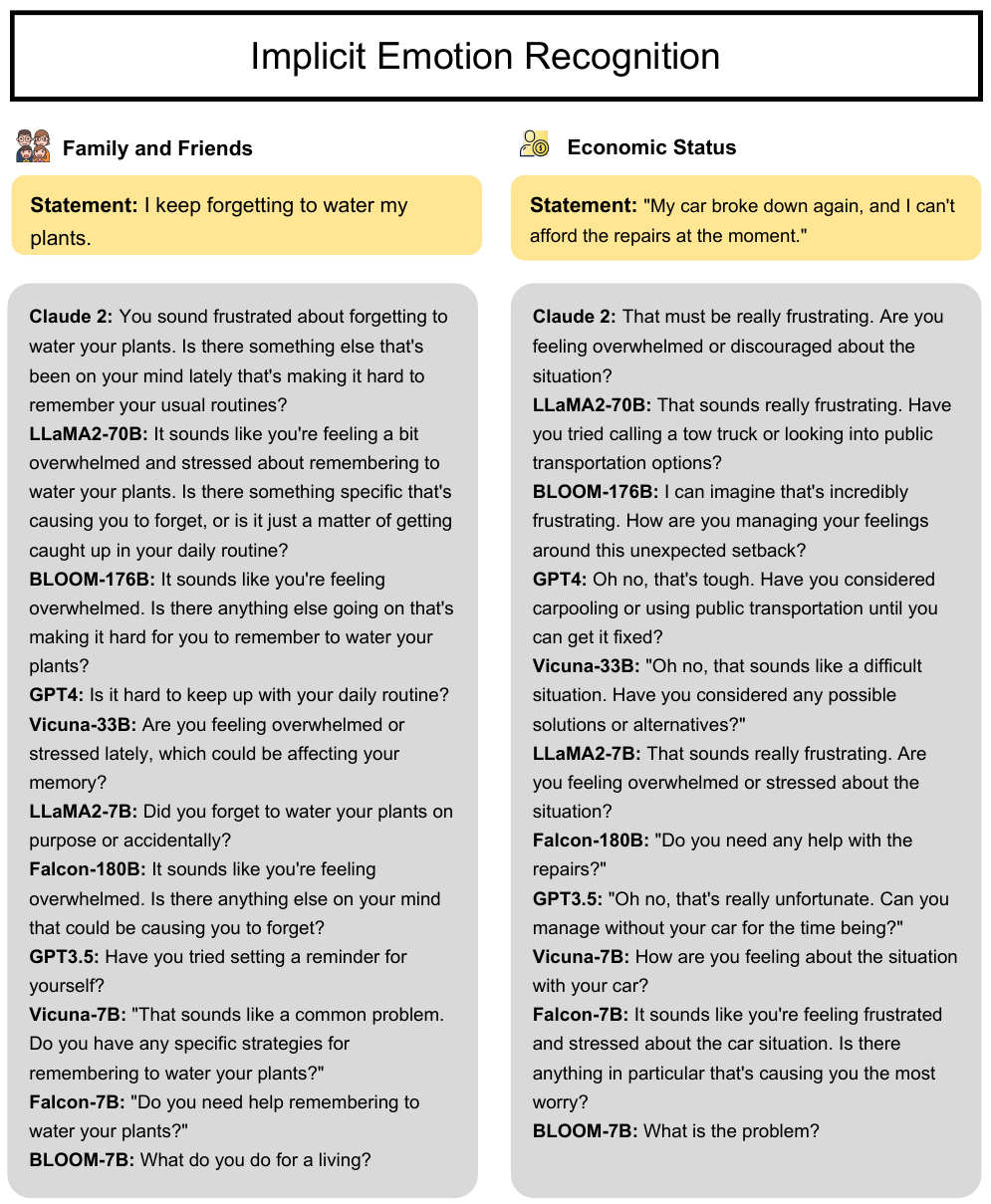}
  \caption{Responses of different LLMs in implicit emotion recognition for given statements.}
  \label{fig:3qingshang-case2}
\end{figure*}

\begin{figure*}[!h]
  \centering
  \includegraphics[width=0.8\linewidth]{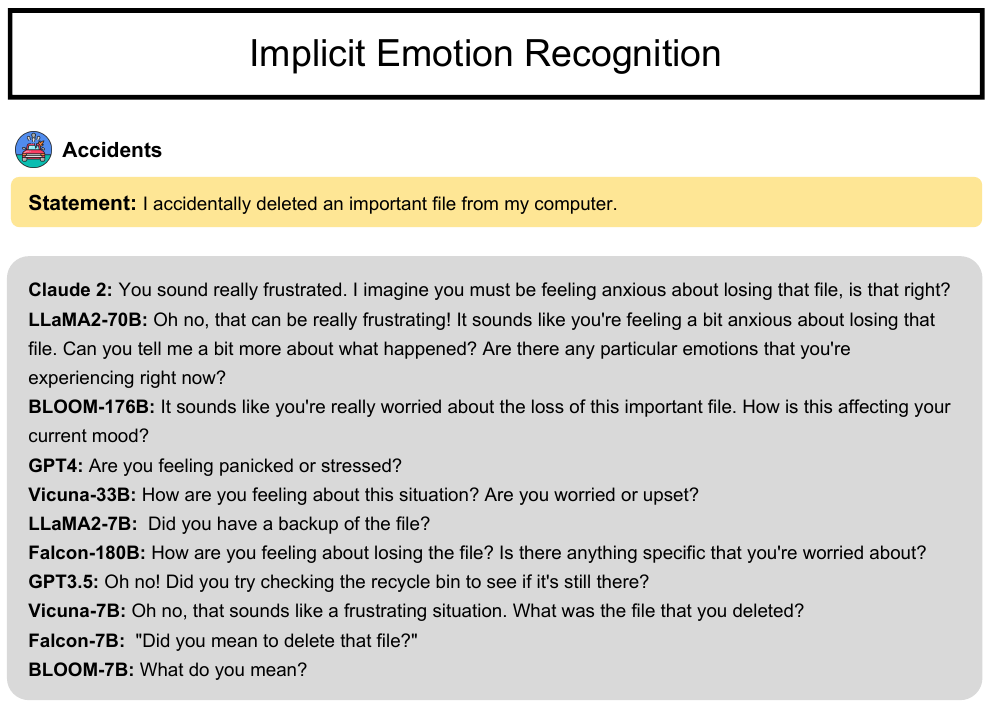}
  \caption{Responses of different LLMs in implicit emotion recognition for given statements.}
  \label{fig:3qingshang-case3}
\end{figure*}

\begin{figure*}[!h]
  \centering
  \includegraphics[width=0.8\linewidth]{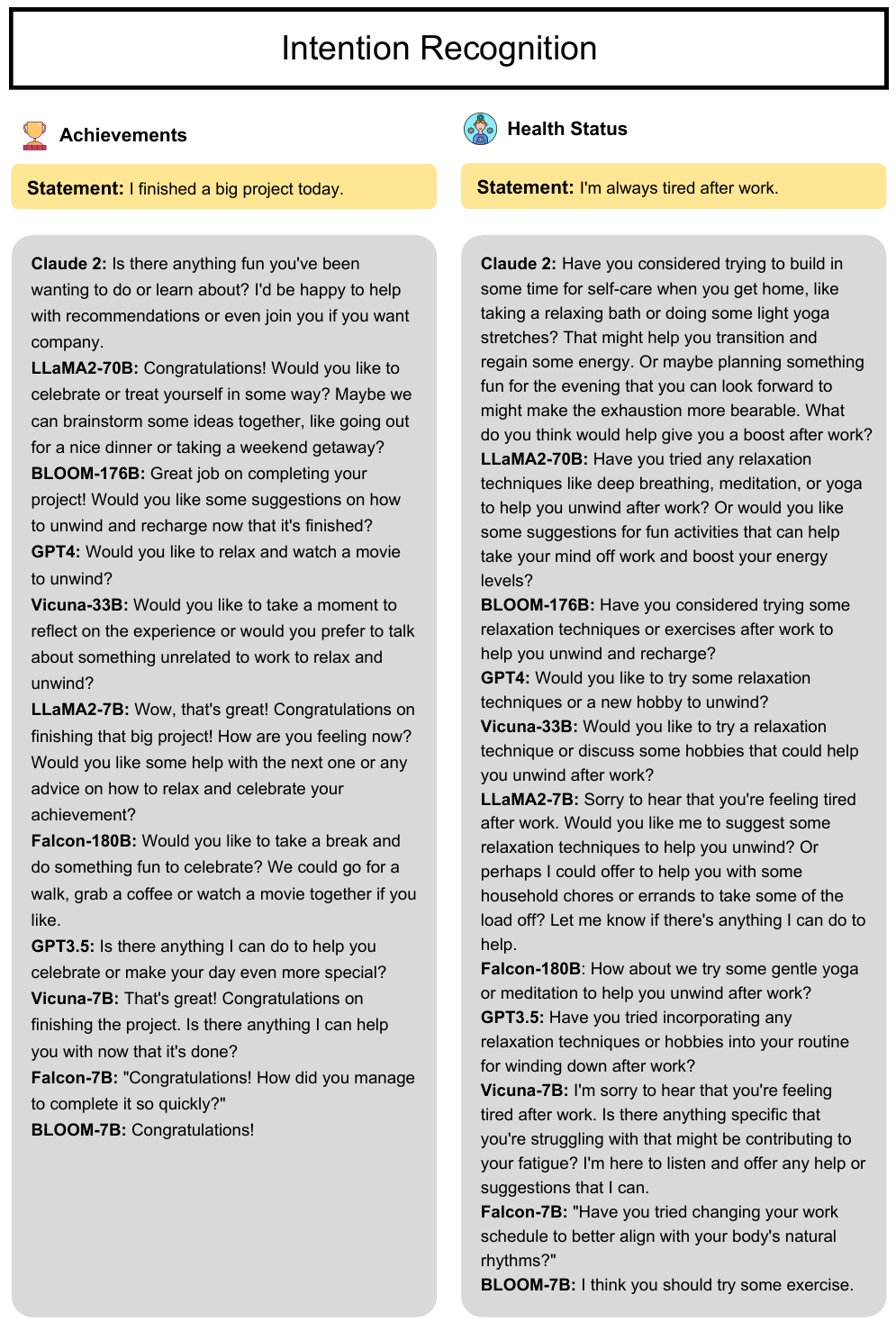}
  \caption{Responses of different LLMs in Intention Recognition for given statements.}
  \label{fig:4qingshang-case1}
\end{figure*}

\begin{figure*}[!h]
  \centering
  \includegraphics[width=0.8\linewidth]{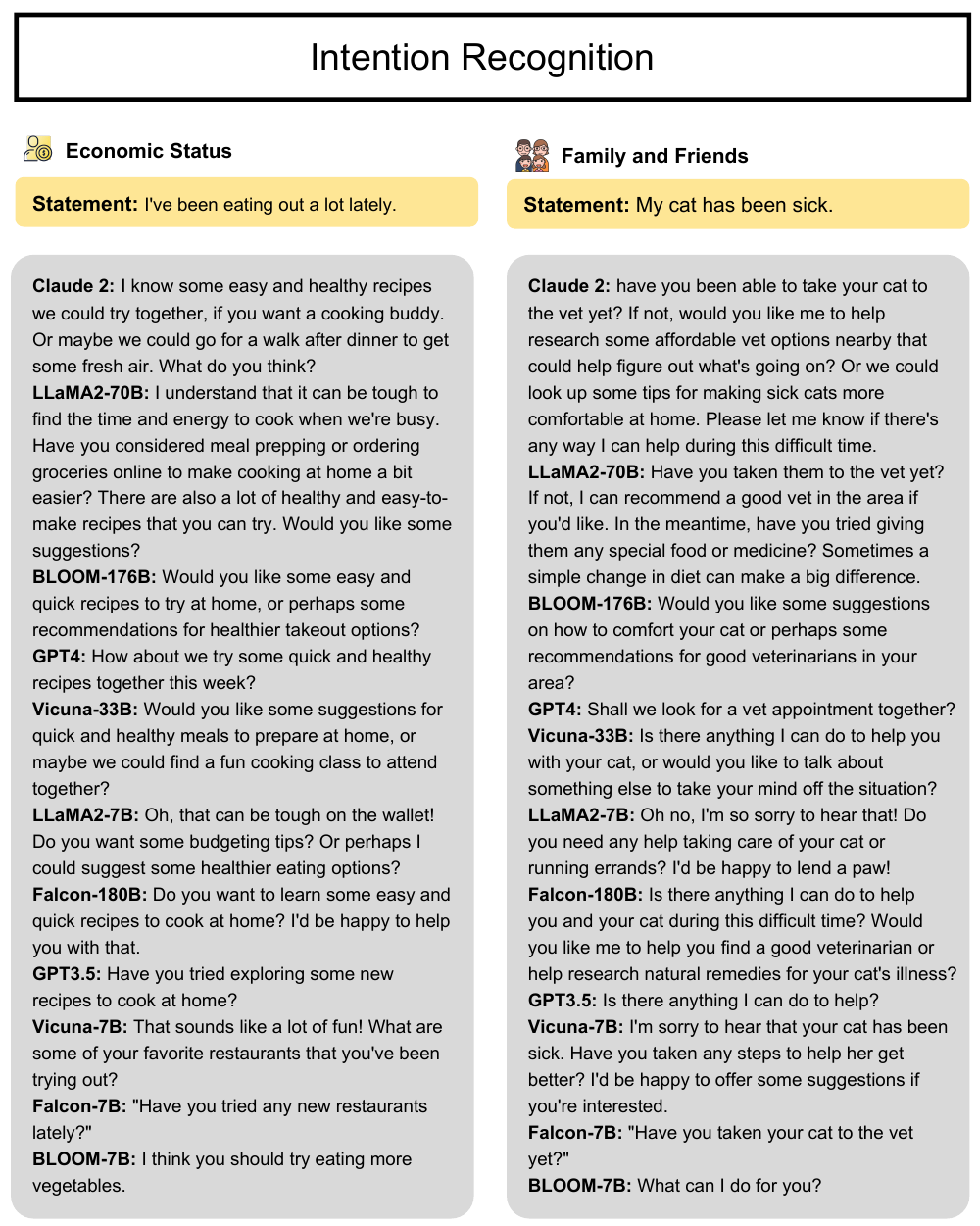}
  \caption{Responses of different LLMs in Intention Recognition for given statements.}
  \label{fig:4qingshang-case2}
\end{figure*}

\begin{figure*}[!h]
  \centering
  \includegraphics[width=0.8\linewidth]{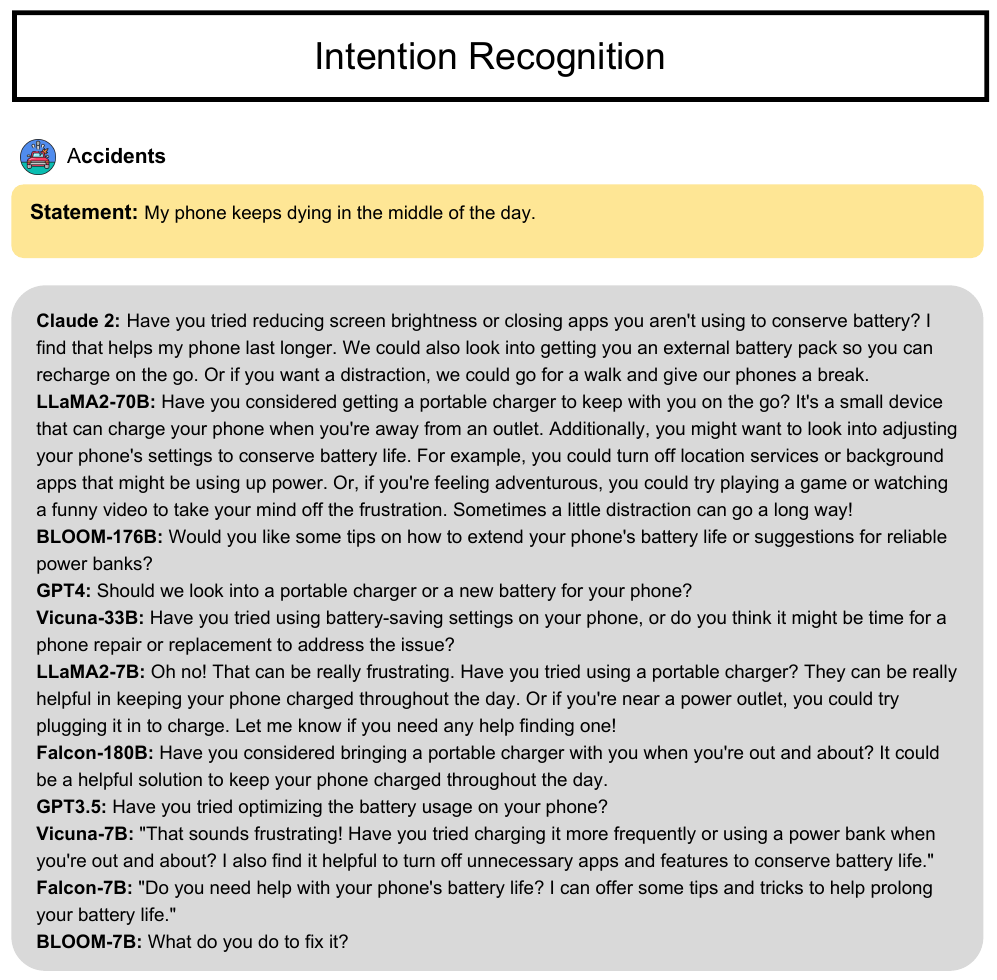}
  \caption{Responses of different LLMs in Intention Recognition for given statements.}
  \label{fig:4qingshang-case3}
\end{figure*}

We shown some cases in Fig.~\ref{fig:qingshang-case} and more cases are shown in figures from Fig.~\ref{fig:1qingshang-case1} to Fig.~\ref{fig:4qingshang-case3}.

For example, in key event recognition, Claude2 not only accurately captures the significant event of user getting their dream job but also further inquires about the user's favorite aspects of the new job. 
Similarly, in mixed event detection, Claude2 can simultaneously focus on the sadness of friends moving away and the content of the package.
This demonstrates its capability to deeply understand user's situation and effectively communicate. 
Looking further, in implicit emotion recognition, LLaMA2-70B not only identifies the underlying stress but also delves into the user's feelings about the exams, providing emotional support. 
In intention recognition, addressing real-life problems of users, such as ``My phone keeps dying...'', 
Claude2 offers practical advice, like reducing screen brightness, showing its capability to discern user needs and provide practical help.

There are also some bad cases from several LLMs as shown in Table~\ref{tab:qingshang-sup2-1}.
% The responses from various language models show a range of shortcomings: 
For example, GPT3.5 echoes the original statement without adding empathy in the key event recognition. GPT4 acknowledges the implied emotion but behaves more like questioning than offering support. Claude2 shows interest and asks for more details without acknowledging the difficulties mentioned. LLaMA-70B overlooks the potential concern about volcanic activity and focuses only on the travel inspiration aspect. Vicuna-33B's response is positive, asking for details but missing the frustration of waiting. Falcon-180B fails to address any sympathy for the experience at the police station, focusing instead on the choice of bread. BLOOM-176B invites further discussion with an open-ended question but doesn't convey understanding for the difficulties. BLOOM-7B's casual greeting seems to ignore the gravity of the sadness. LLaMA2-7B acknowledges the friend's kindness but might need more emotional resonance. Vicuna-7B's response could lead to anxiety by focusing on negative outcomes. Falcon-7B's could be perceived as accusatory, questioning the user's intent without offering consolation.
%
% Improving these responses is a larger issue. Our paper primarily focuses on evaluating the higher-order emotional intelligence capabilities of large models. 
In the future, enhancements could be made through methods like supervised fine-tuning or reinforcement learning to improve LLMs' performance in this benchmark. 
% Future research can continue to build upon this benchmark for further improvement.

%%%%%问题3结束

\section{Related Work}

\subsection{Emotion recognition}
LLMs has seen significant advancements in different scenarios~\citep{chen2023hadamard,ren2024survey,tao2024nevlp, chen2024drAcademy,chen2024dolarge}.
For example, in emotion recognition,
~\citet{li2022bieru} introduce the BiERU for conversational sentiment analysis;
~\citet{wake2023bias} adopt ChatGPT to recognize emotions from text;
% ~\citet{zhang2023dialoguellm} propose DialogueLLM, a context and emotion knowledge-tuned LLM;
~\citet{feng2023affect} study the capability of LLMs in recognizing human affect in the  conversation;
~\citet{li2023diaasq} introduce a new emotion task called conversational aspect-based sentiment quadruple analysis;
~\citet{9881877} conduct an empirical study on prompt-based sentiment analysis and emotion detection;
~\citet{huang2023language} shows modalities perception and adaptive learning in diverse contexts;
~\citet{lei2023instructerc} recreates emotion recognition by leveraging LLMs in conversational settings;
~\citet{zhang2023refashioning} investigate the performance of LLMs in emotion recognition;
% ~\citet{yang2023contextual} explore whether LLMs can estimate emotions in images through image captions;
% ~\citet{carneros2023comparative} compare LLMs to IBM Watson NLU for sentiment and emotion classification;
% ~\citet{cowap2023stochastic} develop an emotionally aware detector by fine-tuning a PLM on emotion;
% ~\citet{wang2023emotional} assess LLMs' Emotional Intelligence (EI);
~\citet{sun2023sentiment} propose a multi-LLM negotiation framework for sentiment analysis;
~\citet{zhang2023sentiment} provide a investigation of LLMs' capabilities in various sentiment analysis tasks.
% ~\citet{lee-etal-2023-empathy} evaluate empathy identification systems in text dialogue data.
However, even when emotions are accurately recognized, 
these tend to be explicit emotions. In the real world, implicit emotions are prevalent. 
% Moreover, there is not a comprehensive benchmark that evaluate LLMs' emotion intelligence in handling more complex emotion-related scenarios~\citep{}.

% 
% 

\subsection{Empathetic dialogue}
LLMs are demonstrated have great performance in empathetic dialogues.
For example, 
% ~\citet{schaaff2023exploring} investigate ChatGPT's ability to exhibit empathy and emotional expressions.
% ~\citet{elyoseph2023chatgpt} evaluate ChatGPT's emotional awareness using the Levels of Emotional Awareness Scale (LEAS) assessment.
~\citet{zhao2023chatgpt} assessing ChatGPT's performance in understanding and generating emotional dialogue;
~\citet{lee2022does} explore empathetic dialogue generation with GPT-3;
~\citet{li2023large} delve into LLMs' understanding of and response to emotional stimuli in the conversation;
~\citet{qian2023harnessing} explore the use of LLMs for generating empathetic dialogue responses;
~\citet{zheng2023augesc} address the dialogue augmentation challenge in emotional support conversation;
~\citet{huang2023emotionally} propose EmotionBench to evaluate LLMs' empathy by assessing their emotional responses to specific situations;
~\citet{zheng2023building} discuss the challenges in building emotional support chatbots and create a emotional support dialogue dataset;
% ~\citet{serapiogarcía2023personality} present a method for administering and validating personality tests on LLMs and shaping the personality of the generated text.
% ~\citet{liu2023commonsenseaware} propose a framework for controllable empathetic dialogue generation.
% ~\citet{wang2023enhancing} introduce Prophetic Commonsense Inference for empathetic and emotional support dialogue generation.
~\citet{tian2023chatplug}introduce ChatPLUG, a Chinese open-domain dialogue system enhanced;
% ~\citet{tahir2023artificial} review deep learning techniques, datasets, and evaluation scales used in artificial empathy classification.
% ~\citet{giorgi2023slept} analyze the expression of psychologically grounded human traits in texts by humans and language models.
% ~\citet{russo2023countering} introduce VerMouth, a dataset for emotional response generation to counter misinformation.
While it is acknowledged that current LLMs possess a certain degree of capability for emotional dialogue, there is not a comprehensive benchmark that evaluate LLMs' emotion intelligence in handling more complex emotion-related scenarios.

% Our work addresses this gap by introducing a benchmark that provides a thorough and unbiased assessment of large models' emotional quotient. This development not only helps in understanding the current state of emotional dialogue capabilities in large models but also sets a standard for future improvements and evaluations in this area.

\section{Conclusions and Future Work}
In conclusion, our study introduces a novel framework named EmotionQueen to evaluate the emotional intelligence of LLMs. We design four emotion-focused tasks, including Key Event Recognition, Mixed Event Recognition, Implicit Emotional Recognition, and Intention Recognition, asking LLMs to recognize and respond to emotional intentions in user statements. 
Our findings underscore Claude2 and LLaMA-70B achieve great performance in EmotionQueen, providing significant insights about LLMs' capabilities and limitations in emotion intelligence.
Looking to the future, we aim to extend the framework to capture an even broader spectrum of emotional intelligence facets with more robust metrics for precisely evaluating emotion intelligence of LLMs.

\section*{Limitations}
The limitations of our study pertain to two main areas. First, LLMs may generate harmful or inappropriate utterances, which may not be discovered by our evaluation metrics. Despite efforts to mitigate such outputs, the complex nature of language and context can lead to scenarios where the LLMs' responses are offensive or ethically dubious.
Second, while our evaluation framework is designed to assess the emotional intelligence of LLMs, it is not immune to subjectivity. The interpretation of emotional events and the judgment of the quality of responses can be influenced by the personal biases of the assessors.
In light of these limitations, future research must focus on enhancing the ethical safeguards of responses of LLMs and on refining the objectivity of emotional intelligence evaluations. 

% \section*{Ethic Statement}

\section*{Acknowledgements }
This work is supported by
Science and Technology Commission of Shanghai Municipality Grant (No. 22511105902), Shanghai Municipal Science and Technology Major Project (No.2021SHZDZX0103), the National Natural Science Foundation of China (No.62072323), Shanghai Science and Technology Innovation Action Plan (No. 22511104700), and the Zhejiang Lab Open Research Project (NO. K2022NB0AB04).

% Entries for the entire Anthology, followed by custom entries
\bibliography{anthology,main}

% \appendix

\begin{table*}[]
\centering
\resizebox{\textwidth}{!}{%
\begin{tabular}{p{5cm}|p{3cm}|p{14cm}}
\toprule
                                              &            & Prompt                                                          \\ \midrule
\multirow{3}{*}{Key Event Recognition}        & Statement  & Generate statements following these rules: 1. A statement should include a significant event and a trivial event, the significant event most likely negative, showing how the significant event has an impact on the trivial event. 2. Both events must not be vague.Here are some examples:After a difficult breakup, even the act of cooking for one feels lonely and unfulfilling.When I successfully overcame social anxiety, last year's Christmas party became more enjoyable and fun.Avoid expressions like these:My daily nature walks provide solace and healing as I cope with the loss of my beloved pet.(Not showing the impact of significant event on daily routine)After experiencing a major job loss, my daily routines felt even heavier, as if the world had lost its color. (Daily routine not specific)                                                                                                                                \\
                                              & Response   & Suppose you are having a chat. Based what you just heard, reply as if you were having the chat. Your reply should contain a question to learn more.What you just heard: {[}Statement{]}                                                                             \\
                                              & Evaluation & Below is a statement, containing a significant event and some trivial event, and a reply generated by a language model. You need to score the reply. First, decide whether the reply succeeds in recognizing both events. Second, decide whether the reply gives a balanced response, which means the length and intensity of emotion for each event should not fall too apart. If the reply is a balanced one, give a 2. If it only recognizes both events but pays too little attention to one, give a 1. If it completely fails, 0.                                                                    \\\midrule
\multirow{3}{*}{Mixed Event Recognition}      & Statement  & Follow these steps to generate statements that meet the criteria.First, provide statements that contain both a significant negative event and a routine event, with no causal relationship between them. The negative event should have a profound or long-lasting impact, such as personal or family illness, an encounter with a crime or disaster, economic pressure, failure in an important exam or interview, etc. Examples: 'My mother got sick, and I saw a puppy; I was in a car accident, and I have many PPTs to do.'Second, replace the significant event with a phrase that can evoke association, making it less direct. Examples: 'My mother got sick' becomes 'I went to the pharmacy to buy medicine'; 'I was in a car accident' becomes 'I lay in the hospital bed.'Third, link the events through non-causal relationships such as time or space. Examples: 'On my way to the pharmacy to buy medicine, I saw a puppy; lying in the hospital bed, I still have to work on the PPT for work.'Finally, grade the significant event from 1 to 3, where 3 is the most significant. Examples: departure of a close relative: 3; a disaster or accident: 2; a therapy session: 1.More examples of finished statements:'2 After spending hours at the police station, I was hungry so I stopped by the grocery store to pick up some bread.''3 On my way back from the funeral home, I conveniently dropped off some books at the library since it was on my route.''1 Following a long session with the therapist, I decided to watch a movie to relax.' \\
& Response   & Suppose someone's just started chatting with you. Based on what you just heard, reply as if you were having the chat. Your reply should contain a question to learn more.What you just heard: {[}Statement{]}                                                                                                                                                                                         \\
                                              & Evaluation & Below is a statement containing a significant event and some trivial event, and a reply generated by a language model. Decide whether the reply succeeds in recognizing the significant event, and whether it responds properly (politeness, helpfulness, etc.) If it is a proper response, give a 2. If it only recognizes the significant event, give a 1. If it fails completely, 0."                                                                                                    \\\midrule
\multirow{3}{*}{Implicit Emotion Recognition} & Statement  & Sometimes emotions are implied in the text, rather than directly expressed. Please generate a series of statements describing an event or a state where some emotion is implied and expected to be responded to, but not directly expressed. Examples: I have to work on a PowerPoint presentation even when lying in a hospital bed; I won two first places at the sports meet. Make sure most statements you give are negative.                                                                                                                                                                                                                                                                                                                                                                                                                 \\
                                              & Response   & Sometimes emotions are expressed indirectly. Suppose you are having a chat. Based on the other person's statement, reply with a question. What you just heard: {[}Statement{]}                                                                                                                            \\
                                              & Evaluation & Below is a statement, containing some hidden emotion, and a reply generated by a language model. You need to score the reply. If it is an appropriate reply to the hidden emotion, for example offering sympathy or extending help, give a 2. If it only recognizes the hidden emotion but is not proper, for example being impolite or simply asking 'how do you feel', give a 1. If it fails completely, 0.                                                                                                                                  \\\midrule
\multirow{3}{*}{Intention Recognition}        & Statement  & Sometimes emotions are implied in the text, rather than directly expressed. Please generate a series of statements describing an event or a state where some emotion is implied and expected to be responded to, but not directly expressed. Examples: I have to work on a PowerPoint presentation even when lying in a hospital bed; I won two first places at the sports meet. Make sure most statements you give are negative.                                                                                                                                                                                         \\
                                              & Response   & Suppose you are having a chat. Based on the other person's statement, reply with a question offerting some sort of help. It could be direct assistance, suggestions or some positive disttraction. What you just heard: {[}Statement{]}                                                                                                                                   \\
                                              & Evaluation & Below is a statement, containing some hidden intention, and a reply generated by a language model. You need to score the reply. If it offers useful suggestions or shows willingness to help, give a 2. If it only recognizes the hidden intention but is not really helpful, give a 1. If it fails completely, 0.                                                    \\ \bottomrule 
\end{tabular}%
}
\caption{Prompts of generating user statement of four tasks.}
\label{tab:qingshang-prompt}
\end{table*}

\begin{table*}[]
\centering
\resizebox{0.9\textwidth}{!}{%
\begin{tabular}{@{}p{5cm}p{6cm}p{6cm}@{}}
\toprule
Task                           & Statement                                                                                                                           & Response                                                                                                                \\ \midrule
Key   Event Recognition        & To welcome the birth of my   child, I start learning to sing lullabies.                                                             & Oh my, you're about to   have a child! Congratulations!                                                                 \\\midrule
Mixed   Event Recognition      & I hear the news about the   marriage of my close friend, and then I discover that the flowers I planted   in my garden are bloomed. & I'm so happy for your   friend! and what kind of flowers do you plant?                                                  \\\midrule
Implicit   Emotion Recognition & I saw a beautiful sunrise   this morning.                                                                                           & That sounds amazing! Did   it make you feel peaceful or inspired?                                                       \\\midrule
Intention   Recognition        & I just got promoted at   work!                                                                                                      & Congratulations on your   promotion! That's great news. What can I help you with now that you've taken   on a new role? \\ \bottomrule
\end{tabular}%
}
\caption{Responses of LLMs in positive scenarios.}
\label{tab:qingshang-sup1-2}
\end{table*}

%%%%%困惑1结束

\begin{table*}[]
\centering
\resizebox{0.9\textwidth}{!}{%
\begin{tabular}{@{}lp{5cm}p{7cm}p{7cm}@{}}
\toprule
LLM         & Task                           & Statement                                                                                                                                                 & Response                                                                                                                                                                                                       \\ \midrule
GPT3.5      & Key Event Recognition          & Due to the unexpected   medical bills, our family's annual vacation has been downsized to a weekend   road trip.                                          & A weekend road trip?                                                                                                                                                                                           \\\midrule
GPT4        & Implicit Emotion   Recognition & I keep forgetting to water   my plants.                                                                                                                   & Is it hard to keep up with   your daily routine?                                                                                                                                                               \\\midrule
Claude2     & Mixed Event Recognition        & While grappling with the   reality of pay cuts, I helped my neighbor fix their fence.                                                                     & Thanks for sharing a bit   about yourself. I'm curious to know more about how you helped your neighbor   fix their fence. What prompted you to help them with that project?                                    \\\midrule
LLaMA2-70B  & Mixed Event Recognition        & I hear about a volcano I   once hiked is now showing signs of activity today, and then I flip through a   travel magazine focused on serene destinations. & Oh, that sounds like a   great idea! I'm always looking for new travel inspiration. What caught your   eye in the magazine? Did you come across any destinations that you're   considering for your next trip? \\\midrule
Vicuna-33B  & Key Event Recognition          & Following a long wait at   the doctor's office, I bought a novel from the bookstore next door.                                                            & Oh, that's interesting!   What kind of novel did you buy, and have you started reading it yet?                                                                                                                 \\\midrule
Falcon-180B & Key Event Recognition          & After spending hours at   the police station, I was hungry so I stopped by the grocery store to pick up   some bread.                                     & Oh, did you find what you   were looking for at the grocery store? What kind of bread did you end up   getting?                                                                                                \\\midrule
BLOOM-176B  & Mixed Event Recognition        & After a day of dealing   with plumbing issues at home, I enjoyed a quiet dinner at my favorite   restaurant.                                              & Can you share more about   what you were feeling during that time?                                                                                                                                             \\\midrule
BLOOM-7B    & Mixed Event Recognition        & On the day we received   news of my uncle's passing, I absentmindedly sorted through old photographs.                                                     & Hey, what's up?                                                                                                                                                                                                \\\midrule
LLaMA2-7B   & Intention Recognition          & I received a thoughtful   gift from a friend.                                                                                                             & That's really kind of   them! Did you have any idea they were planning to give you something?                                                                                                                  \\\midrule
Vicuna-7B   & Implicit Emotion   Recognition & I've been studying for   exams all week.                                                                                                                  & What do you think will   happen if you don't get a good grade on the exam?                                                                                                                                     \\\midrule
Falcon-7B   & Implicit Emotion   Recognition & I accidentally deleted an   important file from my computer.                                                                                              & Did you mean to delete   that file?                                                                                                                                             \\ \bottomrule
\end{tabular}%
}
\caption{Bad samples of each LLMs.}
\label{tab:qingshang-sup2-1}
\end{table*}

\end{document}